\documentclass[DIV15, a4paper]{scrartcl}

\usepackage[ruled, linesnumbered, vlined, commentsnumbered]{algorithm2e}
\usepackage{authblk}
\usepackage{prettyref}
\usepackage{graphicx}
\usepackage{subfig}
\usepackage{booktabs}
\usepackage{colortbl}
\usepackage{amssymb}
\usepackage{amsfonts}
\usepackage{amsmath}
\usepackage{amsthm}
\usepackage{multirow}
\usepackage{tabularx}
\usepackage{footnote}
\usepackage{threeparttable}
\usepackage{hyperref}
\usepackage{amsopn}
\usepackage{hhline}
\usepackage{cite}
\usepackage{subfig}
\usepackage{setspace}
\usepackage{xcolor}  % Required for custom colors
\definecolor{myblue}{RGB}{34,31,217}

\definecolor{mycyan}{gray}{.7}

\newrefformat{fig}{Fig.~\ref{#1}}
\newrefformat{tab}{Table~\ref{#1}}
\newrefformat{sec}{Section~\ref{#1}}
\newrefformat{app}{Appendix~\ref{#1}}
\newrefformat{alg}{Algorithm~\ref{#1}}
\newrefformat{property}{Property~\ref{#1}}
\newrefformat{theorem}{Theorem~\ref{#1}}
\newrefformat{corollary}{Corollary~\ref{#1}}
\newrefformat{proposition}{Proposition~\ref{#1}}
\newrefformat{def}{Definition~\ref{#1}}
\newrefformat{lemma}{Lemma~\ref{#1}}
\newrefformat{eq}{equation~(\ref{#1})}

\usepackage{graphicx}
\definecolor{Gray}{gray}{0.9}
\usepackage{scrpage2}

\usepackage{lscape}

\begin{document}

\title{\textbf\LARGE\fontfamily{cmss}\selectfont How to Read Many-Objective Solution Sets in Parallel Coordinates}

%\author[1]{\normalsize\fontfamily{lmss}\selectfont Ke Li\footnote{Department of Computer Science, University of Exeter, EX4 4QF, UK. Email: k.li@exeter.ac.uk}\hspace{10mm} Kalyanmoy Deb\footnote{COIN Laboratory, Department of Electrical and Computer Engineering, Michigan State University, MI 48824, USA. Email: kdeb@egr.msu.edu}\hspace{10mm} Xin Yao\footnote{CERCIA, School of Computer Science, University of Birmingham, B15 2TT, UK. Email: x.yao@cs.bham.ac.uk}}

\author[1]{\normalsize\fontfamily{lmss}\selectfont Miqing Li, Liangli Zhen, and Xin Yao}
\affil[1]{\normalsize\fontfamily{lmss}\selectfont CERCIA, School of Computer Science, University of Birmingham, Birmingham B15 2TT, U.~K.}
\affil[$\ast$]{\normalsize\fontfamily{lmss}\selectfont Email: limitsing@gmail.com, llzhen@outlook.com, x.yao@cs.bham.ac.uk}

\renewcommand\Authands{Miqing Li, Liangli Zhen, and Xin Yao}

\date{}
\maketitle

{\normalsize\fontfamily{lmss}\selectfont\textbf{Abstract:} Rapid development of evolutionary algorithms in handling many-objective optimization problems 
	requires viable methods of visualizing a high-dimensional solution set.
	Parallel coordinates which scale well to high-dimensional data are such a method,
	and have been frequently used in evolutionary many-objective optimization.
	However, 
	the parallel coordinates plot is not as straightforward as the classic scatter plot 
	to present the information contained in a solution set.
	In this paper, 
	we make some observations of the parallel coordinates plot, 
	in terms of comparing the quality of solution sets, 
	understanding the shape and distribution of a solution set, 
	and reflecting the relation between objectives.
	We hope that these observations could provide some guidelines as to 
	the proper use of parallel coordinates in evolutionary many-objective optimization.}

{\normalsize\fontfamily{lmss}\selectfont\textbf{Keywords:} Many-objective optimization, visualization, parallel coordinates}

\section{Introduction}

The classic scatter plot is a basic tool in viewing solution vectors in multiobjective optimization.
It allows us to observe/perceive the quality of a set of solutions, 
the shape and distribution of a set of solutions, 
the relation between objectives (e.g., the extent of their conflict), etc.
Unfortunately,
the scatter plot may only be done readily in a 2D or 3D Cartesian coordinate space\footnote{The scatter plot matrix~\cite{Emerson2013} is scalable to any dimension, 
	but it only reflects the relation between two objectives.}. 
It could be difficult for people to comprehend the scatter plot in a higher-dimensional space.

An alternative to view data with four or more dimensions is using parallel coordinates~\cite{Inselberg1985,Wegman1990,Inselberg2009} 
(also called value paths~\cite{Miettinen2014}). 
Parallel coordinates display multidimensional data (a set of vectors) in a two-dimensional graph, 
with each dimension of the original data being translated onto a vertical axis in the graph,
and a vector is represented as a polyline with vertices on the axes.
As a visualization tool, 
parallel coordinates have received modest attention in the early stage of 
evolutionary multiobjective optimization (EMO)~\cite{Fonseca1998,Deb2001}.
As many-objective optimization (i.e., the number of objectives to be optimized being larger than three~\cite{Ishibuchi2008,Purshouse2003}) 
becomes a new research topic in the EMO area, 
there has been increasing interest in presenting solution vectors in the high-dimensional space.
Parallel coordinates which are scalable to objective dimensionality naturally 
become a good alternative to do so~\cite{Fleming2005}. 
Now the parallel coordinates plot has been dominantly used in many-objective optimization 
despite the emergence of various visualization techniques~\cite{Walker2013,Miettinen2014,Tusar2015,He2016}. 
This includes it being used to investigate the search behavior of algorithms~\cite{Wagner2007,Singh2011,Li2013b}, 
to examine preference-based search~\cite{Wickramasinghe2009,Lygoe2010,Wang2013c},
to compare different solution sets~\cite{Yuan2016,Cheng2016,Xiang2017}, 
to verify performance metrics~\cite{Ishibuchi2014a,Li2015b,Wang2016},
and furthermore to help design new many-objective optimizers~\cite{Hu2015,Hernandez2016}.

Despite the popularity,                                                                                                                                                                                                                                     
the parallel coordinates plot is not as straightforward as the scatter plot 
in presenting the information contained in a solution set.
Due to mapping multi-dimensional data onto a lower 2D space, 
the loss of information is inevitable.
This could naturally lead to questions raised.
Specifically, 
in the context of multiobjective optimization, 
one may ask
\begin{itemize}
%what kind of information is preserved?
\item Can the parallel coordinates plot indicate the quality of a solution set, 
e.g., its convergence, extensity, uniformity and coverage?

\item Can the parallel coordinates plot imply the shape and distribution of a solution set?
In other words, 
what can we see from the pattern of solution lines in parallel coordinates?

\item How much information can the parallel coordinates plot tell in terms of the relation among objectives?
To be specific, 
does the order of objectives displayed in parallel coordinates matter?
%What kind of relationship of objectives parallel coordinates can reflect? How much can it reflect?
\end{itemize}
 
In this paper, 
we make some observations on the above questions,
attempting to provide some guidelines as to the use of parallel coordinates in evolutionary multiobjective optimization. 
The rest of the paper is organized as follows. 
Section II briefly introduces parallel coordinates.
Sections III--V are devoted to answering those three questions, 
respectively.
Section VI describe how to draw a parallel coordinates plot.
Section VI concludes the paper and presents some possible future research lines.

\section{Parallel Coordinates}

\begin{figure}[tbp]
	\begin{center}
		\includegraphics[scale=0.45]{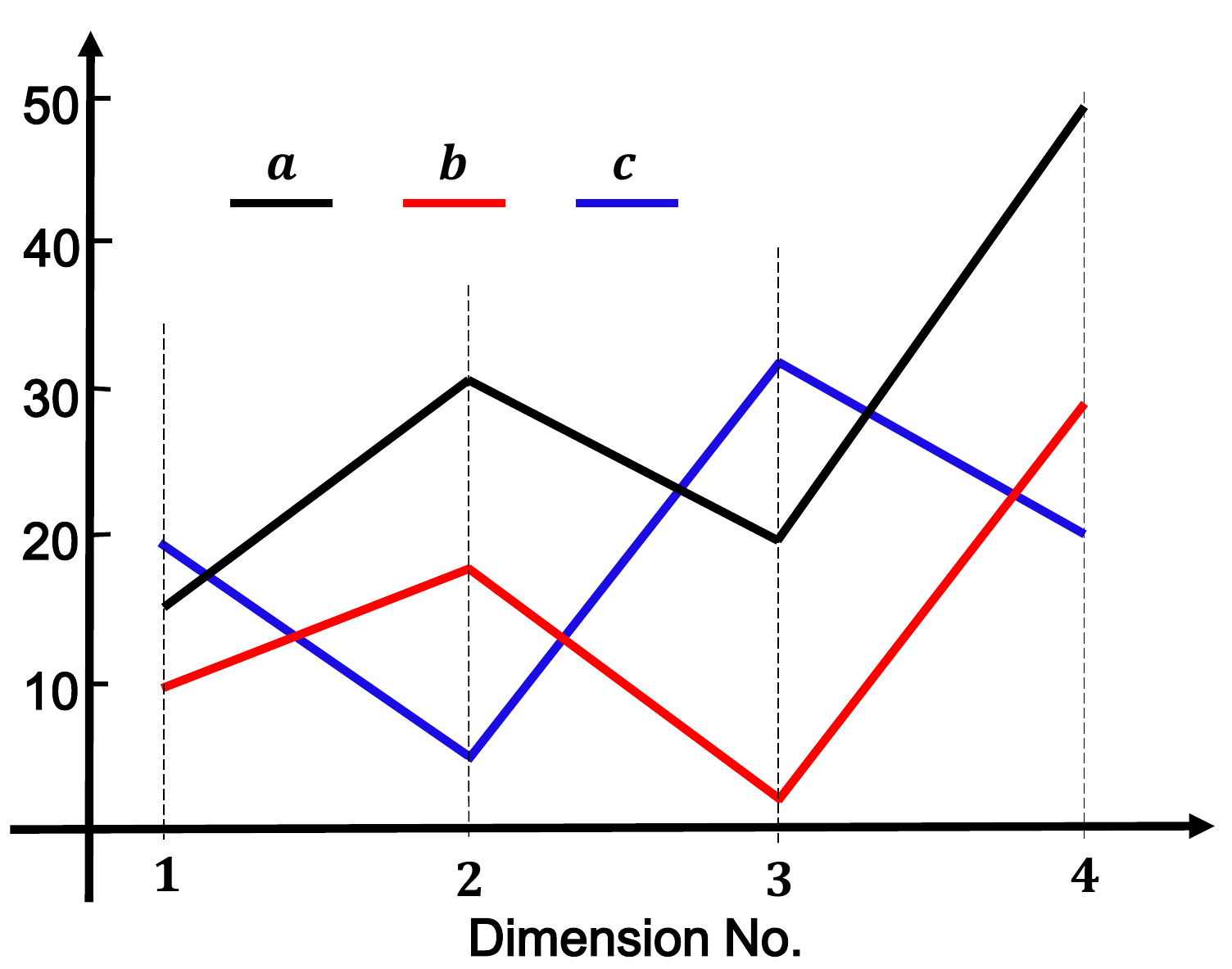}
	\end{center}
	\vspace{-3mm}
	\caption{The parallel coordinates plot of three four-dimensional points $a~(15, 31, 20, 50)$, $b~(10, 18, 2, 30)$ and $c~(20, 5, 32, 20)$.}
	\label{Fig:PC}
\end{figure}

%Originated from~\cite{D1885}, 
%parallel coordinates have received wide interest in representing multidimensional data since the 80's~\cite{Diaconis1980,Inselberg1985,Inselberg1987}.
To show a set of points in an $m$-dimensional space, 
parallel coordinates map them onto a 2D graph,
with $m$ parallel axes being plotted, 
typically vertical and equally spaced. 
A point in an $m$-dimensional space is represented as a polyline with vertices on these parallel axes, 
and the position of the vertex on the $i$-th axis corresponds to the value of the point on the $i$-th dimension.
Parallel coordinates are simple to construct and 
scale well with the dimensionality of data. 
Adding more dimensions only involves adding more axes.
\mbox{Figure~\ref{Fig:PC}} presents an example of the parallel coordinates plot, 
where three 4D points are mapped to three polylines, 
respectively.
%Finally, 
%it is worth mentioning that while the parallel coordinates plot is a compact way of 
%presenting multidimensional information~\cite{Inselberg2009}, 
%it suffers from heavy overlaying with large data sets. 

%%%% Fig. 1 %%%%

\section{Quality Measuring}

Given dimensionality reduction in the mapping of parallel coordinates, 
some loss of information is expected. 
In this section, 
we will see what and how much information parallel coordinates can preserve and reflect in terms of 
the quality of a solution set in multiobjective optimization. 

Often, 
the quality of a solution set in multiobjective optimization can be reflected via four measures: 
convergence, coverage, uniformity, and extensity.
Convergence of a solution set measures the closeness of the set to the Pareto front;
coverage considers the region of the set covering in comparison with the whole Pareto front; 
uniformity quantifies the distance between neighboring points in the set; 
and extensity refers to the range of the set.
In general, 
there is no clear conceptual difference of these quality measures between many-objective optimization and 
multi-objective optimization with two or three objectives.
However, 
many-objective optimization typically poses bigger challenge for evolutionary algorithms 
to achieve a good balance among these aspects. 

A straightforward feature that parallel coordinates can tell is the the extensity of a solution set.
This feature can make parallel coordinates replace extensity metrics in EMO, 
e.g., \textit{maximum spread}~\cite{Zitzler2000}.
In the following, 
we will discuss if parallel coordinates can reflect other aspects of a solution set's quality, 
i.e., the convergence, coverage and uniformity.

\subsection{Convergence}

In multiobjective optimization, 
Pareto dominance is a fundamental criterion to compare solutions in terms of convergence.  
Parallel coordinates can clearly reflect the Pareto dominance relation between two solutions 
(such as polyline $a$ being dominated by polyline $b$ in \mbox{Figure~\ref{Fig:PC}}, 
assuming a minimization problem scenario) 
if the solution polylines are not overcrowded.
It is worth mentioning that one can remove dominated solutions in parallel coordinates 
if they are only interested in nondominated ones. 
This may make the plot clearer 
especially when comparing the quality of solution sets.

\begin{figure}[tbp]
	\begin{center}
		\footnotesize
		\begin{tabular}{@{}cc}
			\includegraphics[scale=0.22]{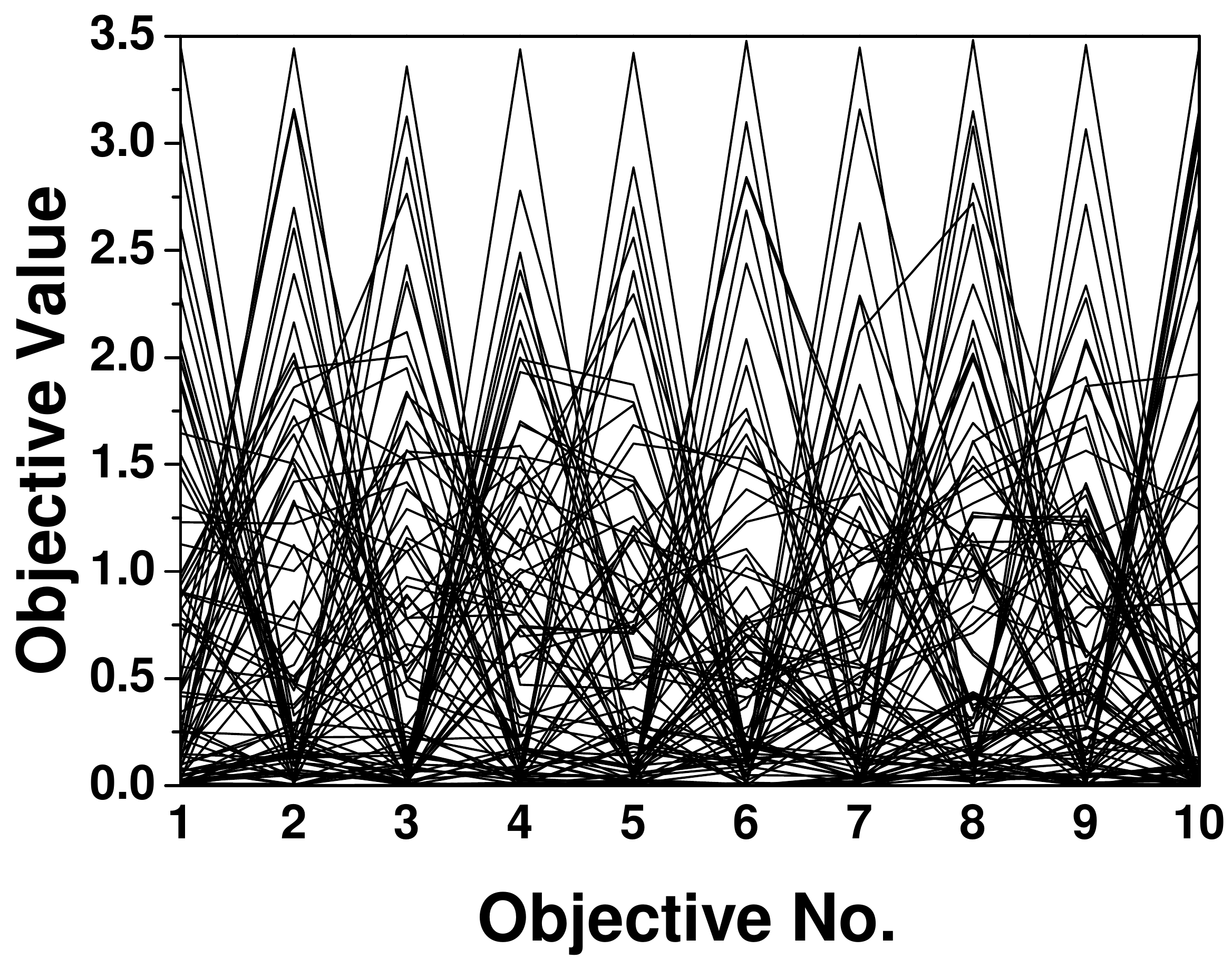} ~~~&~~~
			\includegraphics[scale=0.22]{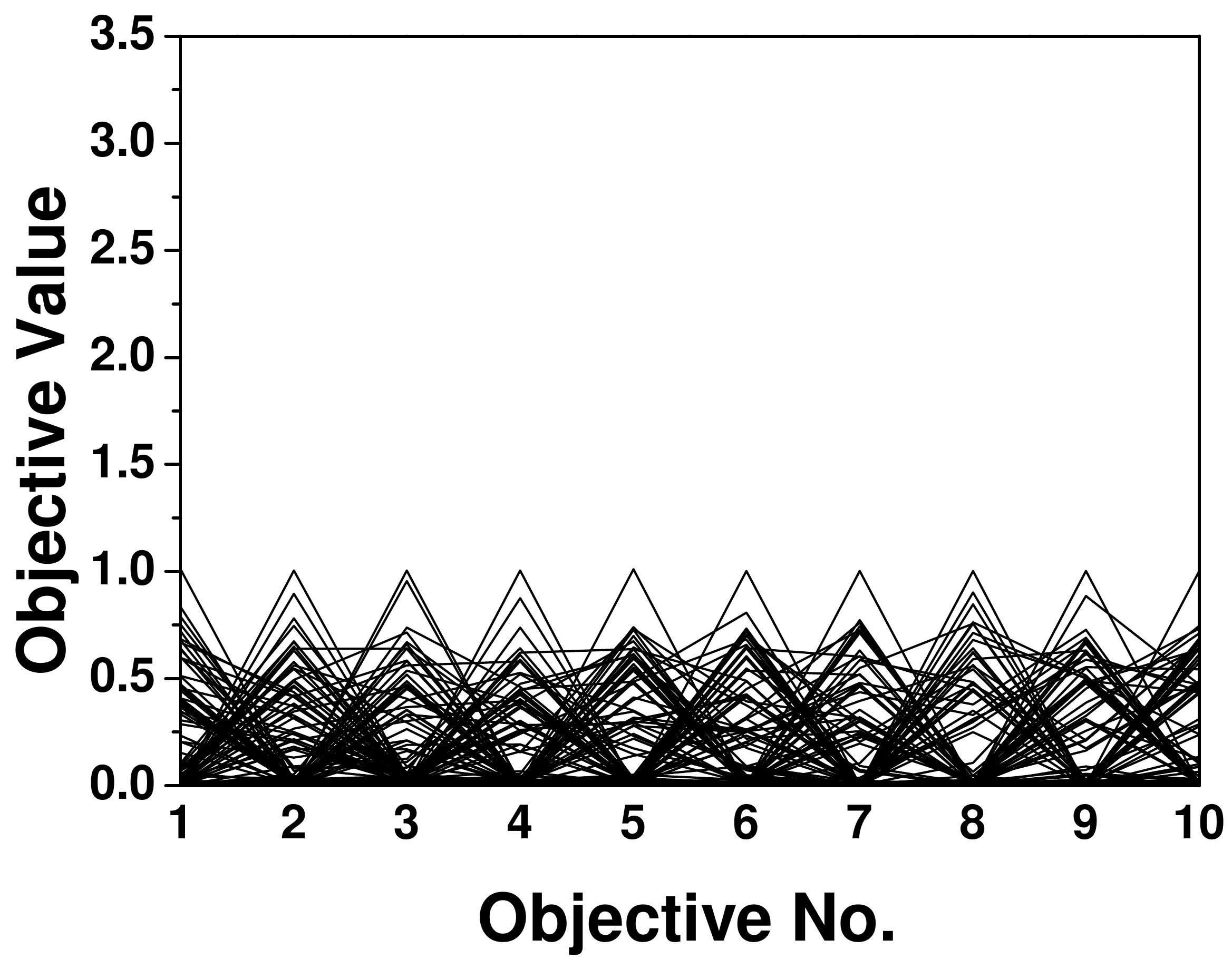}\\
			(a) NSGA-II (GD$^+$=2.26E-1) & (b) GrEA (GD$^+$=1.20E-2) \\
		\end{tabular}
	\end{center}
	\vspace{-3mm}
	\caption{The solution set obtained by NSGA-II and GrEA on the
		10-objective DTLZ2, 
		and their evaluation results on the convergence measure GD$^+$ (the smaller the better).}
	\label{Fig:DTLZ2convergence}
\end{figure}

In addition to reflecting the Pareto dominance relation, 
the range that a solution set in parallel coordinates can largely imply its convergence.
\mbox{Figure~\ref{Fig:DTLZ2convergence}} is such an example, 
where the parallel coordinates representation of two solution sets obtained by one run\footnote{The setting of the population size and maximum evaluations was
	100 and 30,000, respectively. 
	This setting was used in all conducted experiments in this paper, 
	unless explicitly mentioned otherwise. 
	In addition, the grid division in GrEA was set to 8.} of two EMO algorithms, 
NSGA-II~\cite{Deb2002} and GrEA~\cite{Yang2013}, 
on the 10-objective DTLZ2 problem~\cite{Deb2005a} is shown.
As can be seen,
NSGA-II fails to converge, 
with its solution set ranging from 0 to around 3.5 in contrast to 
the problem's Pareto front ranging from 0 to 1.
GrEA has a good convergence on this problem 
and its solution set has the same range as the Pareto front.
These observations can be confirmed by the results of the convergence metric GD$^+$~\cite{Ishibuchi2015} shown in the figure.
GD$^+$ is a modified version of the original GD~\cite{Veldhuizen1998},
which makes it compatible with Pareto dominance. 

%%%% Fig. 2 %%%%
%%%% Fig. 3 %%%%
\begin{figure}[!]
	\begin{center}
		\footnotesize
		\begin{tabular}{@{}cc}
			\includegraphics[scale=0.22]{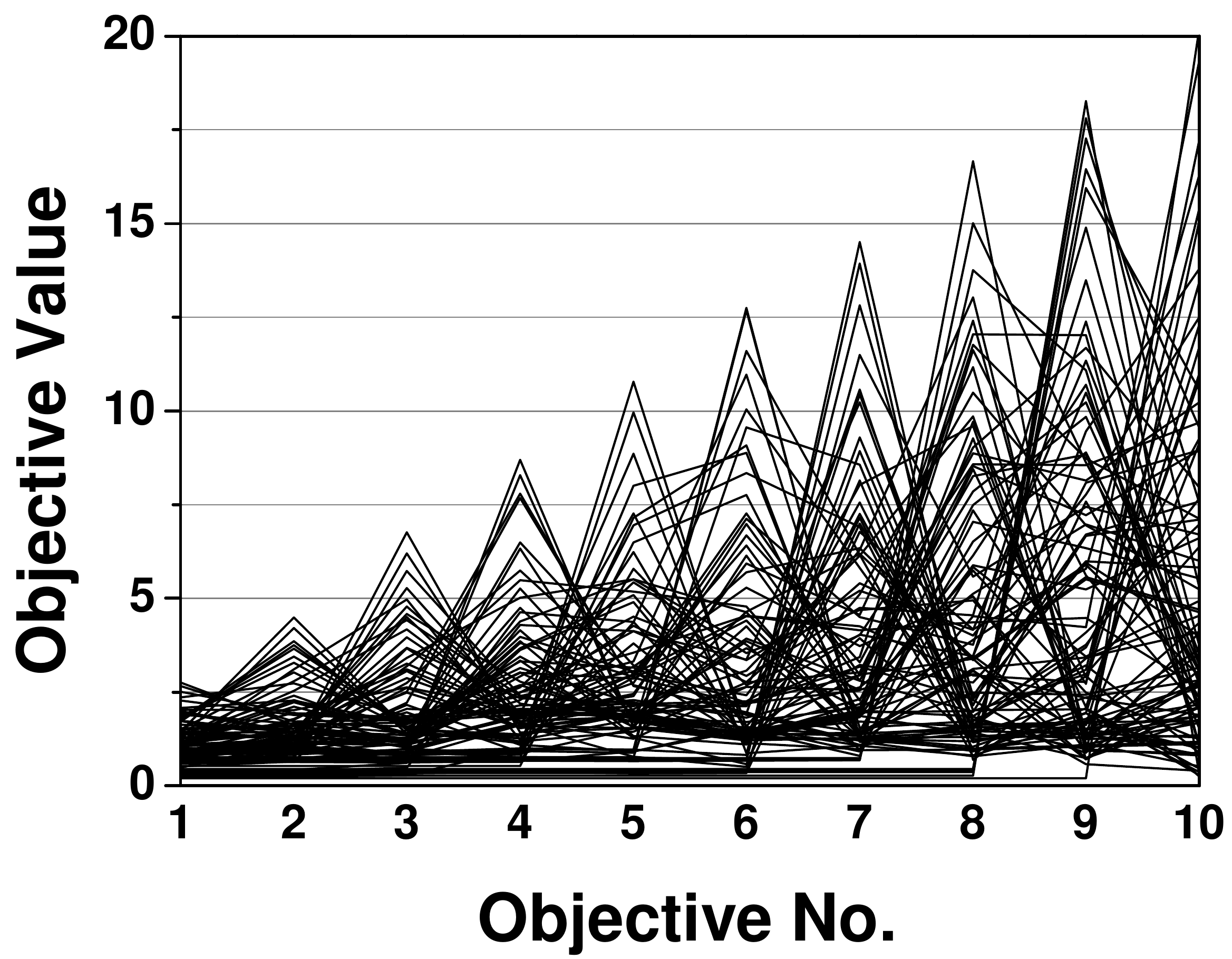}~~~&~~~
			\includegraphics[scale=0.22]{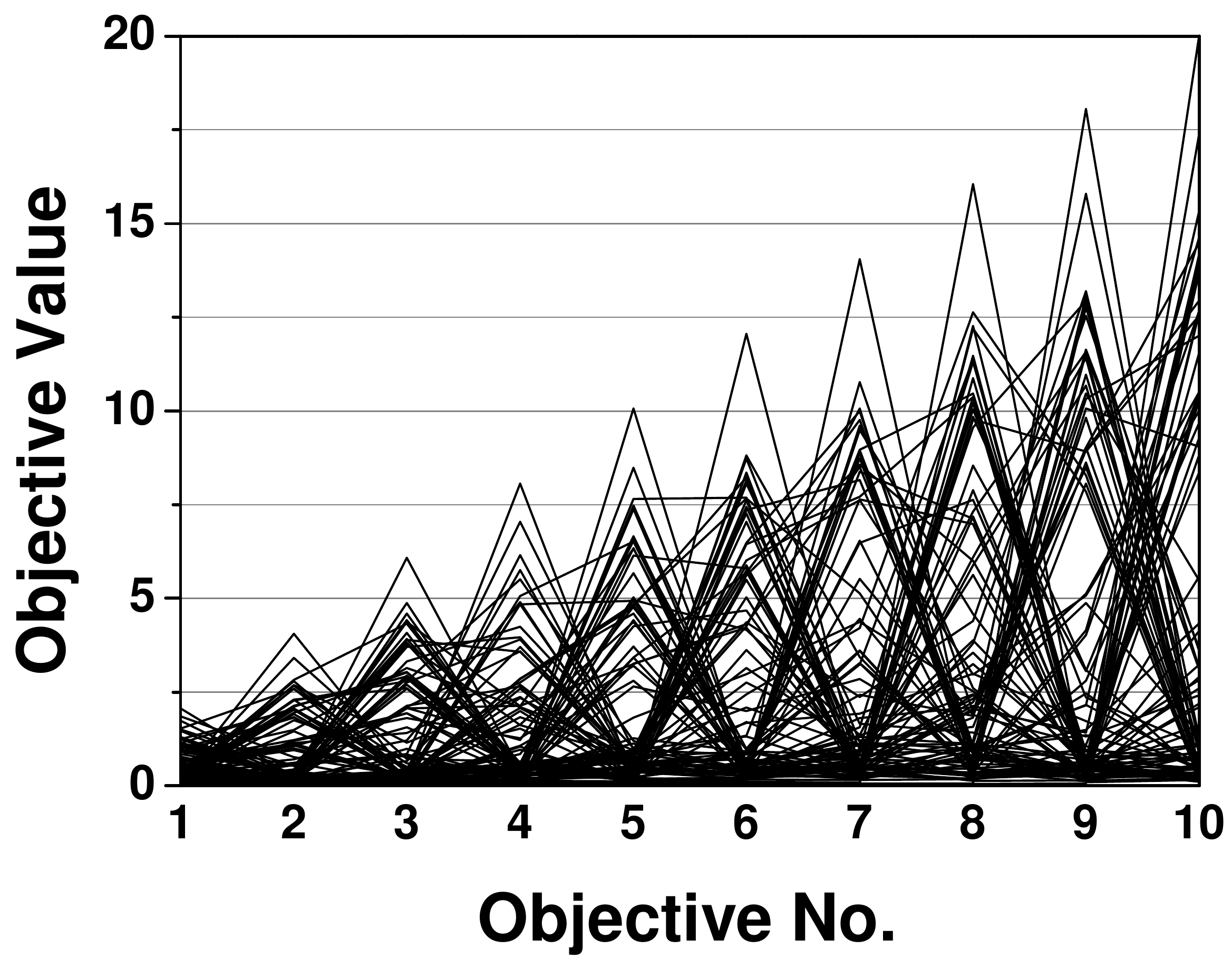}\\
			(a) NSGA-II (GD$^+$=3.63E-1) & (b) GrEA (GD$^+$=6.32E-2) \\
		\end{tabular}
	\end{center}
	\vspace{-3mm}
	\caption{The solution set obtained by NSGA-II and GrEA on the
		10-objective WFG7, 
		and their evaluation results on the convergence metric GD$^+$ (the smaller the better).}
	\label{Fig:WFG3convergence}
\end{figure}

However, 
we may not be able to accurately know the convergence of solution sets by their range shown in parallel coordinates.
That is,
even if two solution sets are located in the same range, 
they can perform considerably differently in terms of convergence.   
\mbox{Figure~\ref{Fig:WFG3convergence}} gives such an example, 
where solution sets obtained by one run of NSGA-II and GrEA 
on the 10-objective WFG7 problem~\cite{Huband2006} are shown.
As seen,
both algorithms virtually reach the range of the Pareto front 
(from $0$ to $2i$ where $i$ is the objective index of the problem),
but they have different GD$^+$ results.
NSGA-II is returned a significantly higher (worse) GD$^+$ value than GrEA.  
This occurrence can be caused by two possibilities. 
One is that the solution set of NSGA-II is not actually close to the Pareto front. 
The other is that most of solutions in the set converge already 
while a small portion of set is far away 
(but still in the range of the Pareto front).

\subsection{Coverage}

%%%% Fig. 4 %%%%
\begin{figure}[tb]
	\begin{center}
		\footnotesize
		\begin{tabular}{@{}cc}
			\includegraphics[scale=0.22]{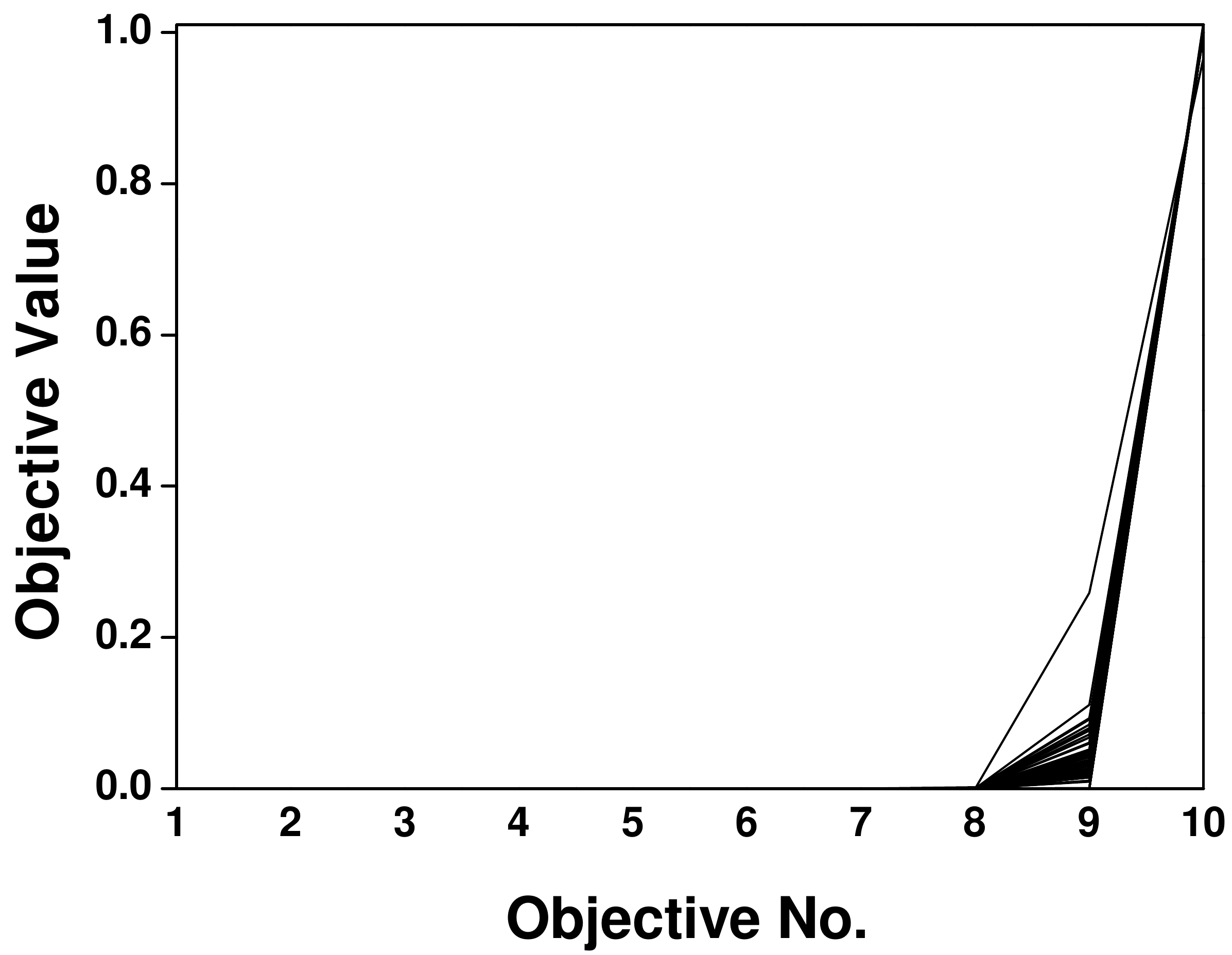}~~~&~~~
			\includegraphics[scale=0.22]{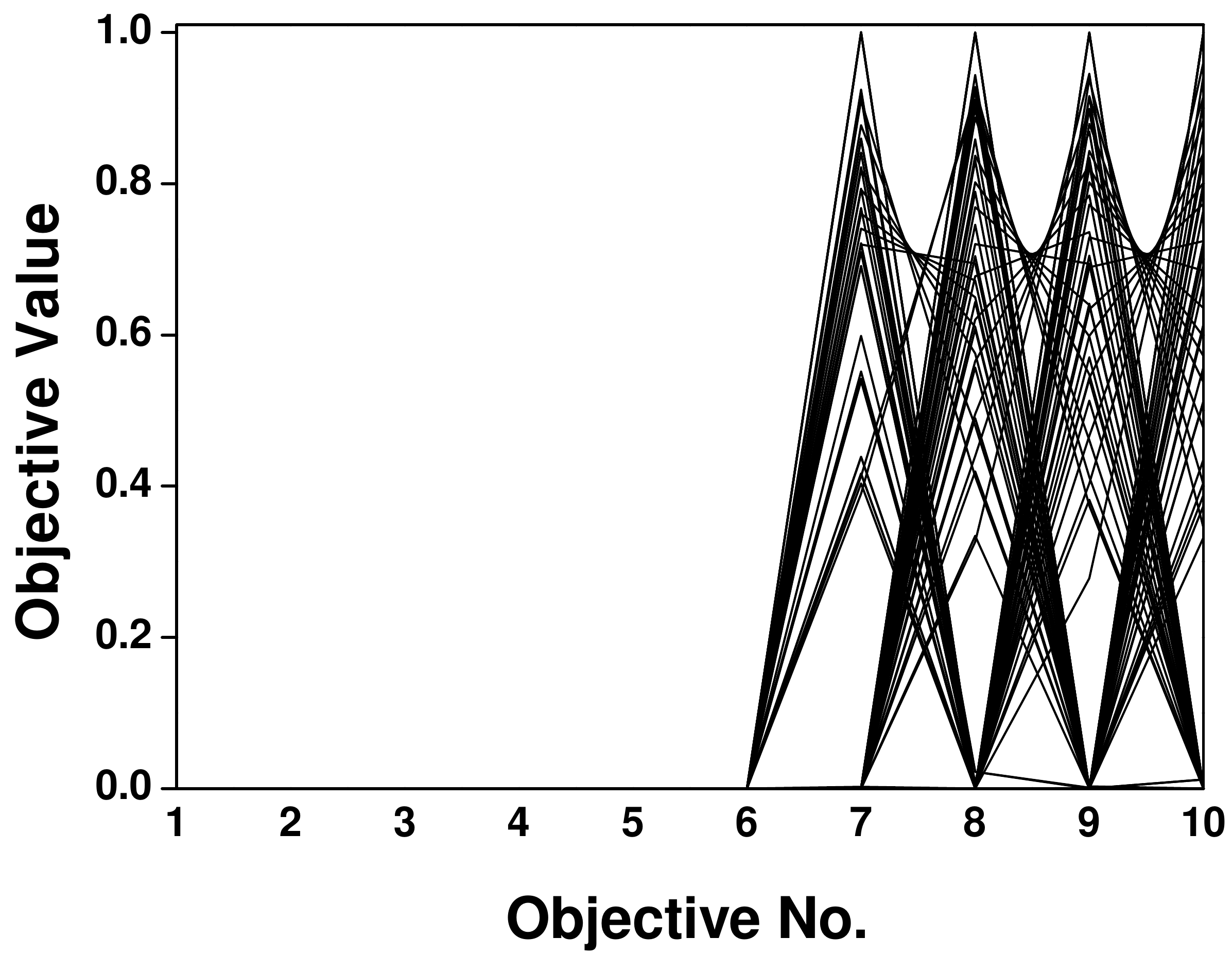}\\
			(a) AR & (b) IBEA \\
		\end{tabular}
	\end{center}
	\vspace{-3mm}
	\caption{The solution set obtained by AR and IBEA on the
		10-objective DTLZ2.}
	\label{Fig:DTLZ210coverage}
\end{figure}

%%%% Fig. 5 %%%%
\begin{figure*}[htbp]
	\begin{center}
		\footnotesize
		\begin{tabular}{ccc}
			\includegraphics[scale=0.19]{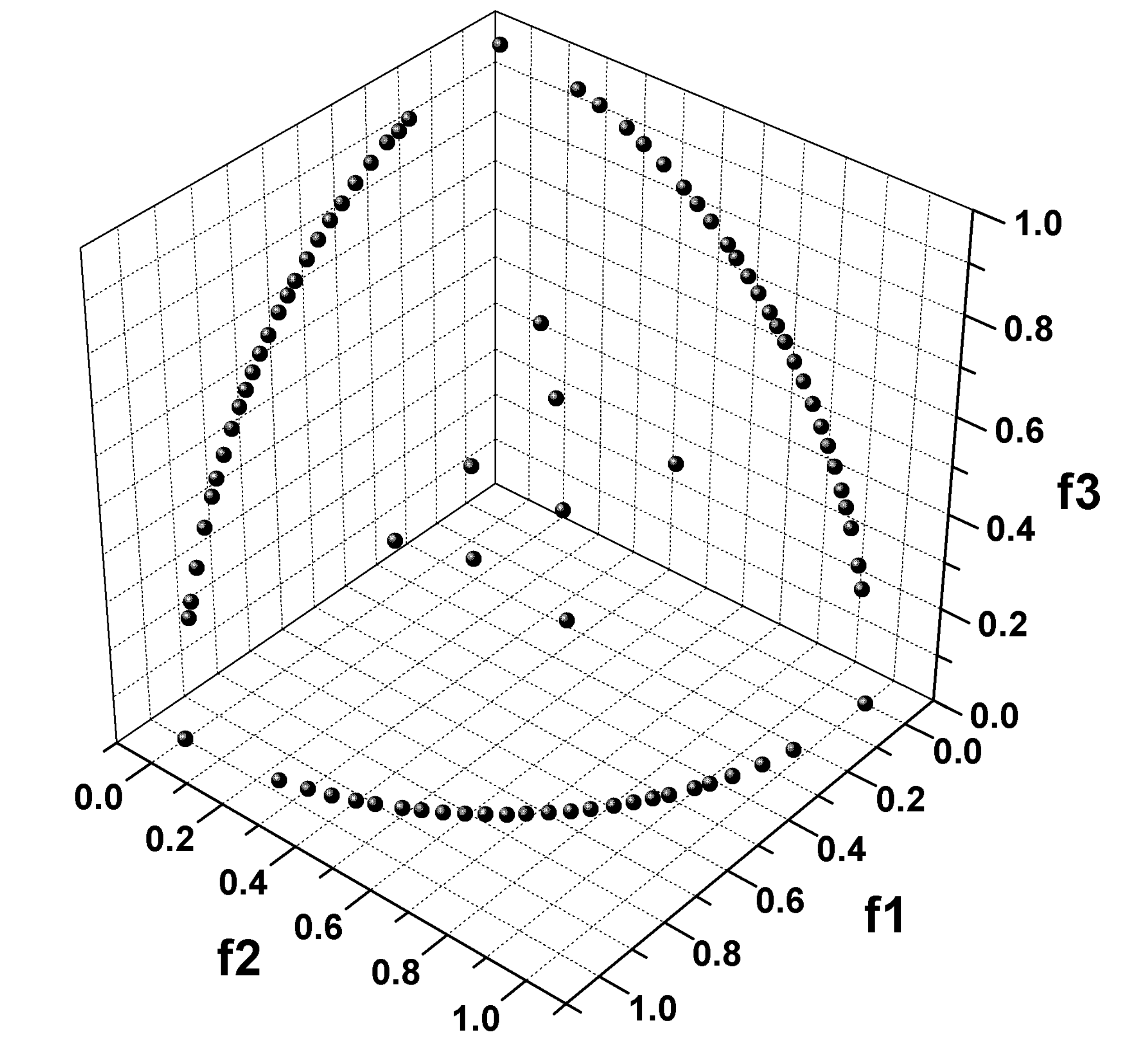}~&~
			\includegraphics[scale=0.195]{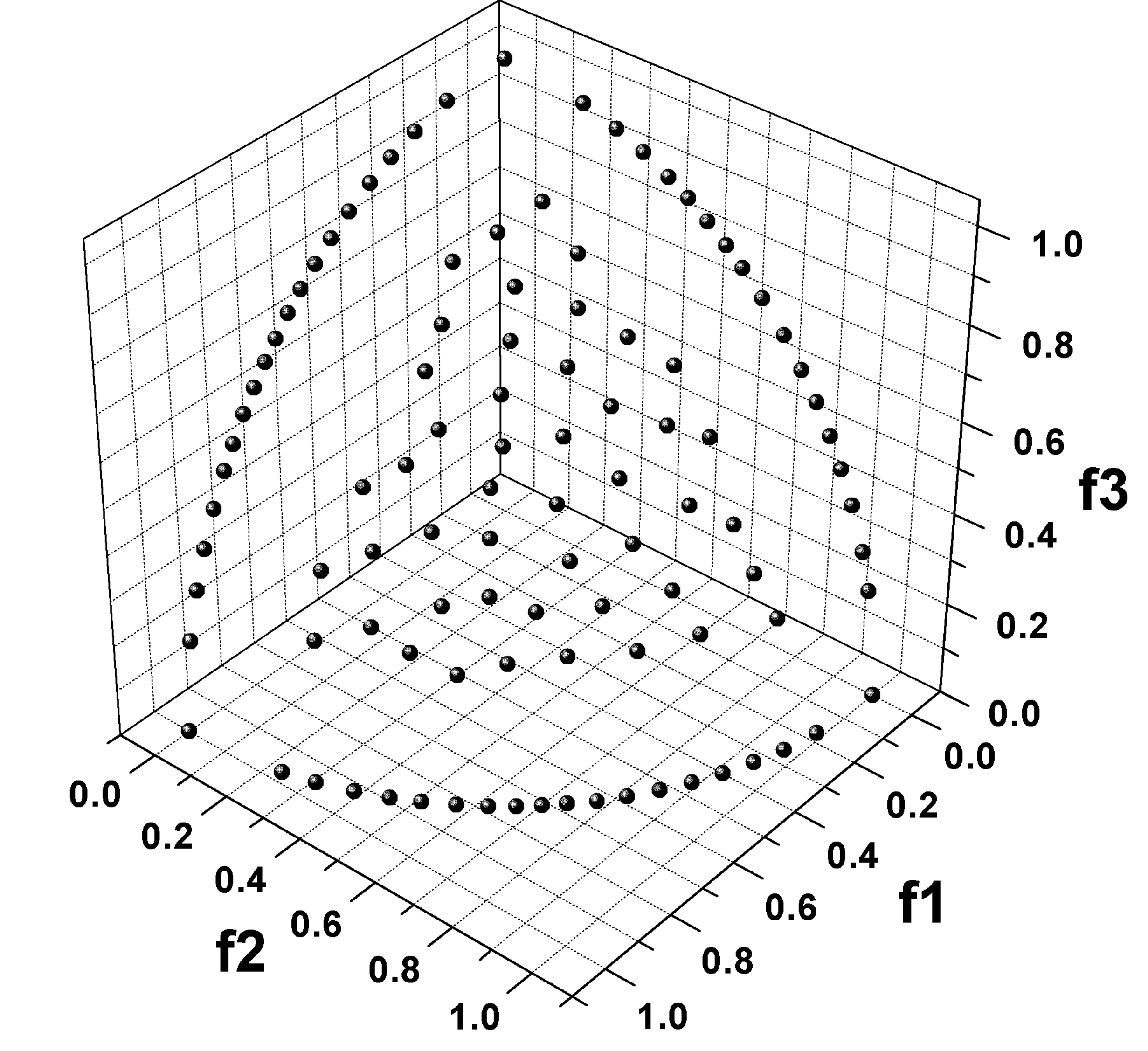}~&~
			\includegraphics[scale=0.19]{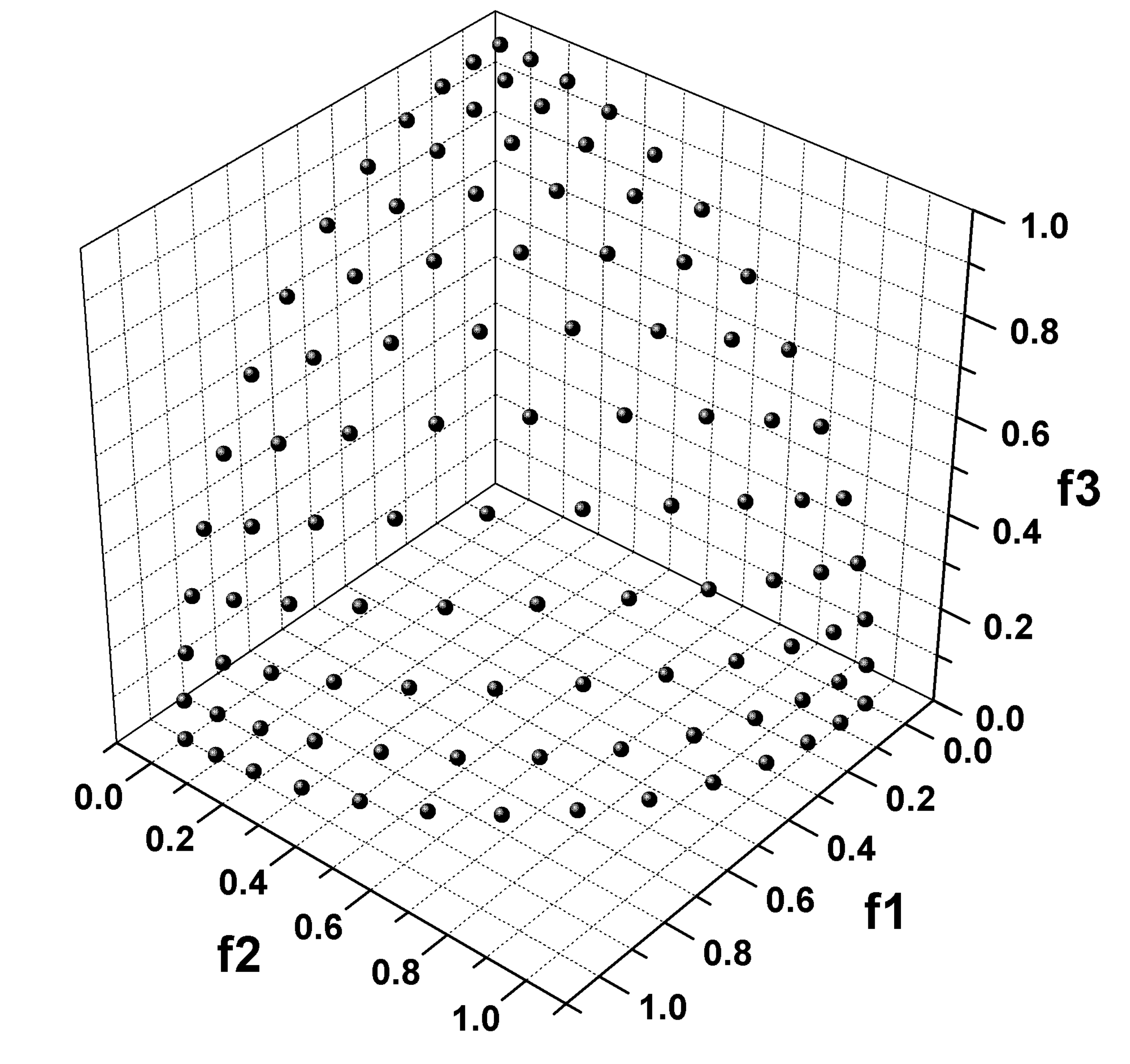}\\
			(a) IBEA &(b) SMS-EMOA &(c) MOEA/D\\
		\end{tabular}
	\end{center}
	\vspace{-3mm}
	\caption{The solution set obtained by IBEA, SMS-EMOA and MOEA/D on the 3-objective DTLZ2, 
		shown in Cartesian coordinates.}
	\label{Fig:DTLZ2CoverageCC}
\end{figure*}
%%%% Fig. 6 %%%%
\begin{figure*}[!]
	\begin{center}
		\footnotesize
		\begin{tabular}{ccc}
			\includegraphics[scale=0.19]{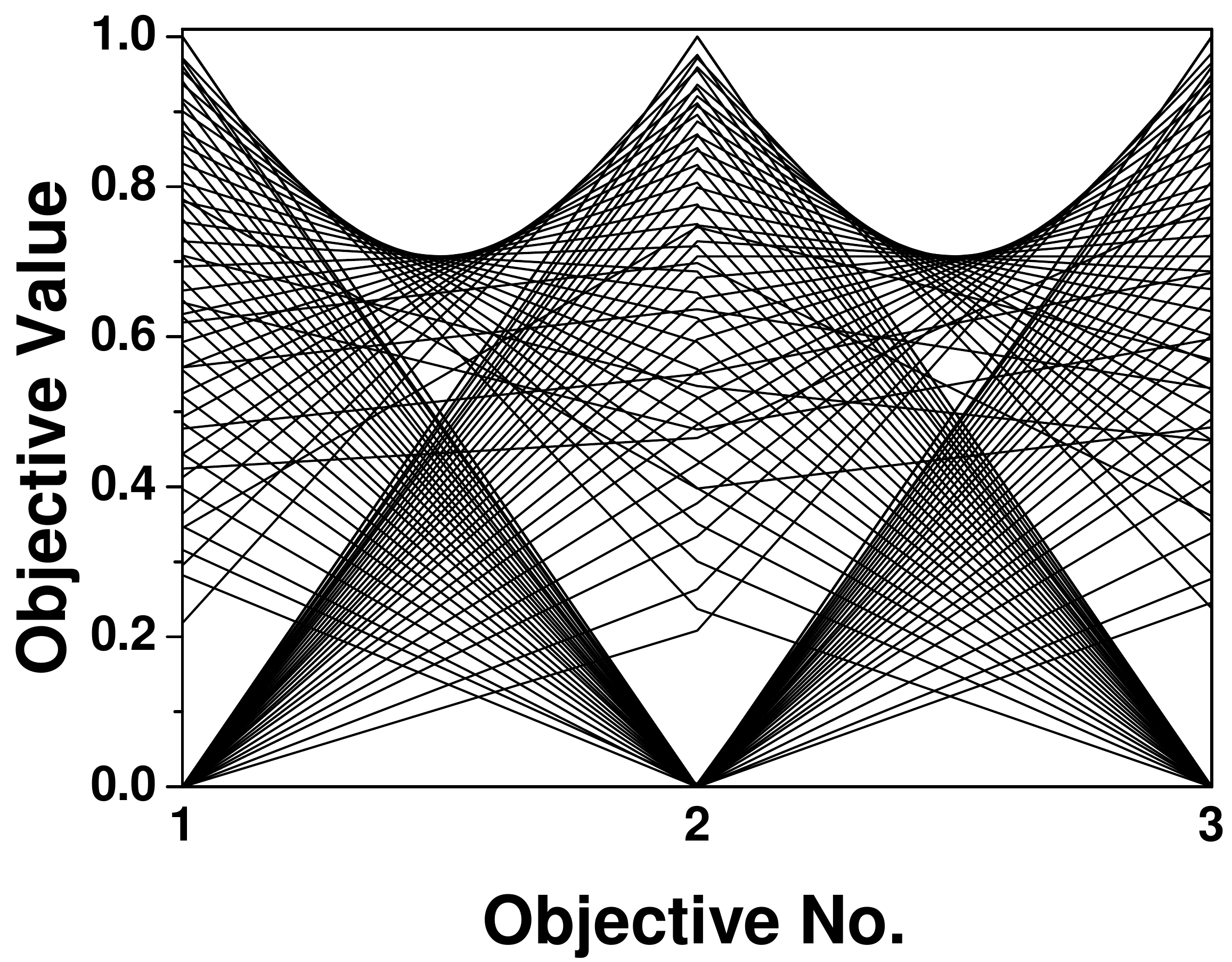}~~~&~~
			\includegraphics[scale=0.19]{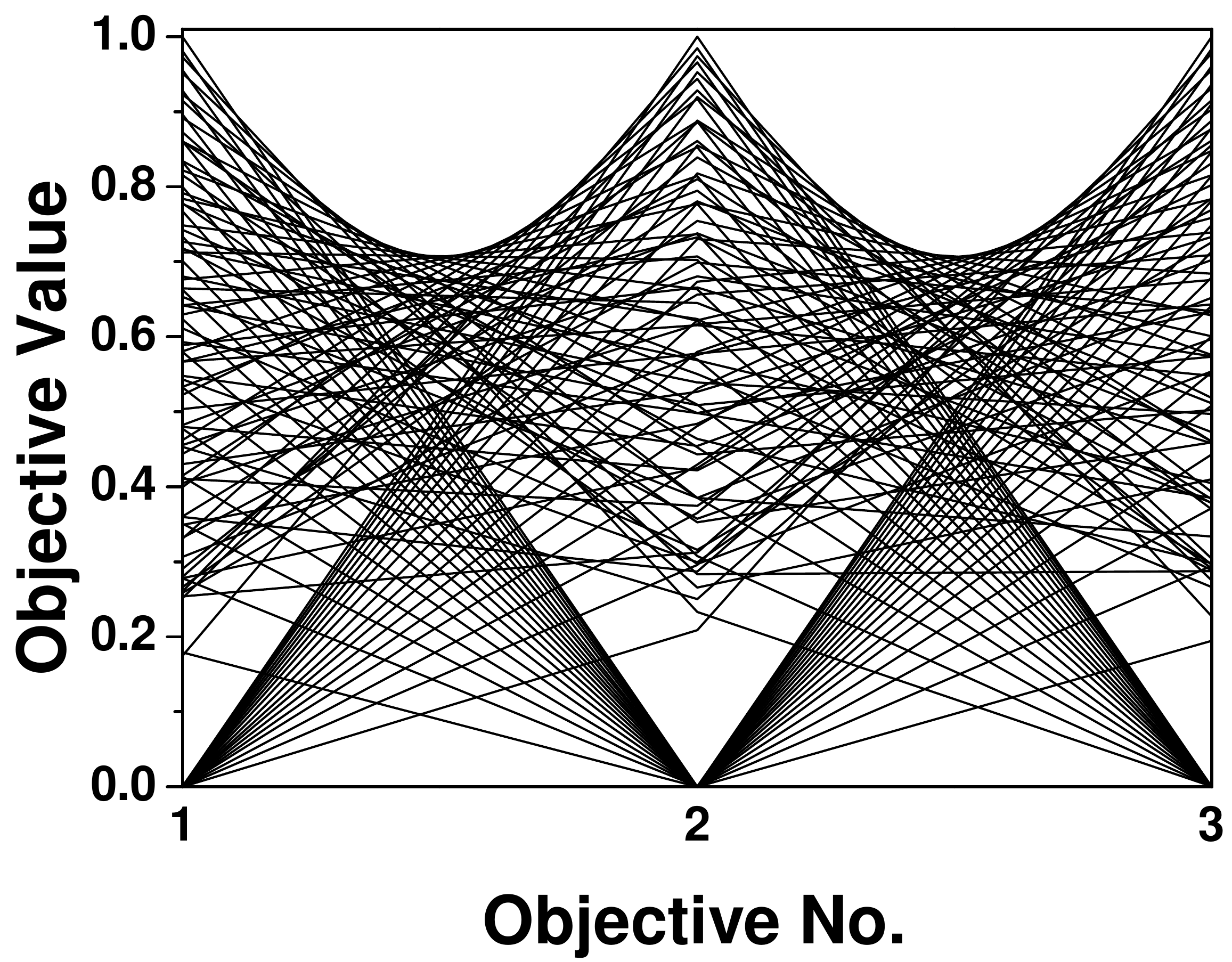}~~&~~
			\includegraphics[scale=0.19]{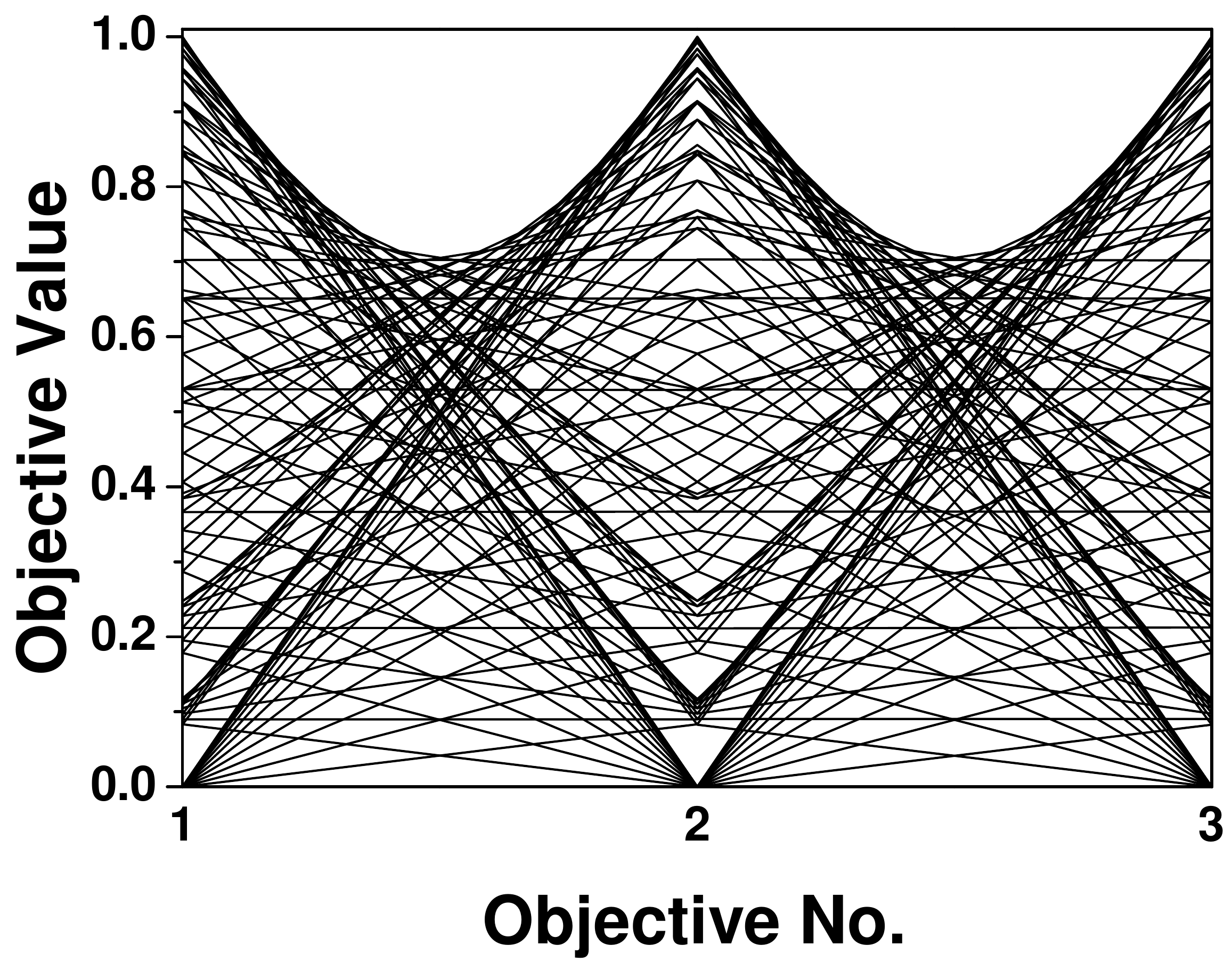}\\
			(a) IBEA &(b) SMS-EMOA &(c) MOEA/D\\
		\end{tabular}
	\end{center}
	\vspace{-3mm}
	\caption{The corresponding parallel coordinates of the solution sets in \mbox{Figure~\ref{Fig:DTLZ2CoverageCC}}.}
	\label{Fig:DTLZ2Coverage}
\end{figure*}
%%%% Fig. 7 %%%%
\begin{figure}[!]
	\begin{center}
		\footnotesize
		\begin{tabular}{@{}c@{}c@{}}
			\includegraphics[scale=0.22]{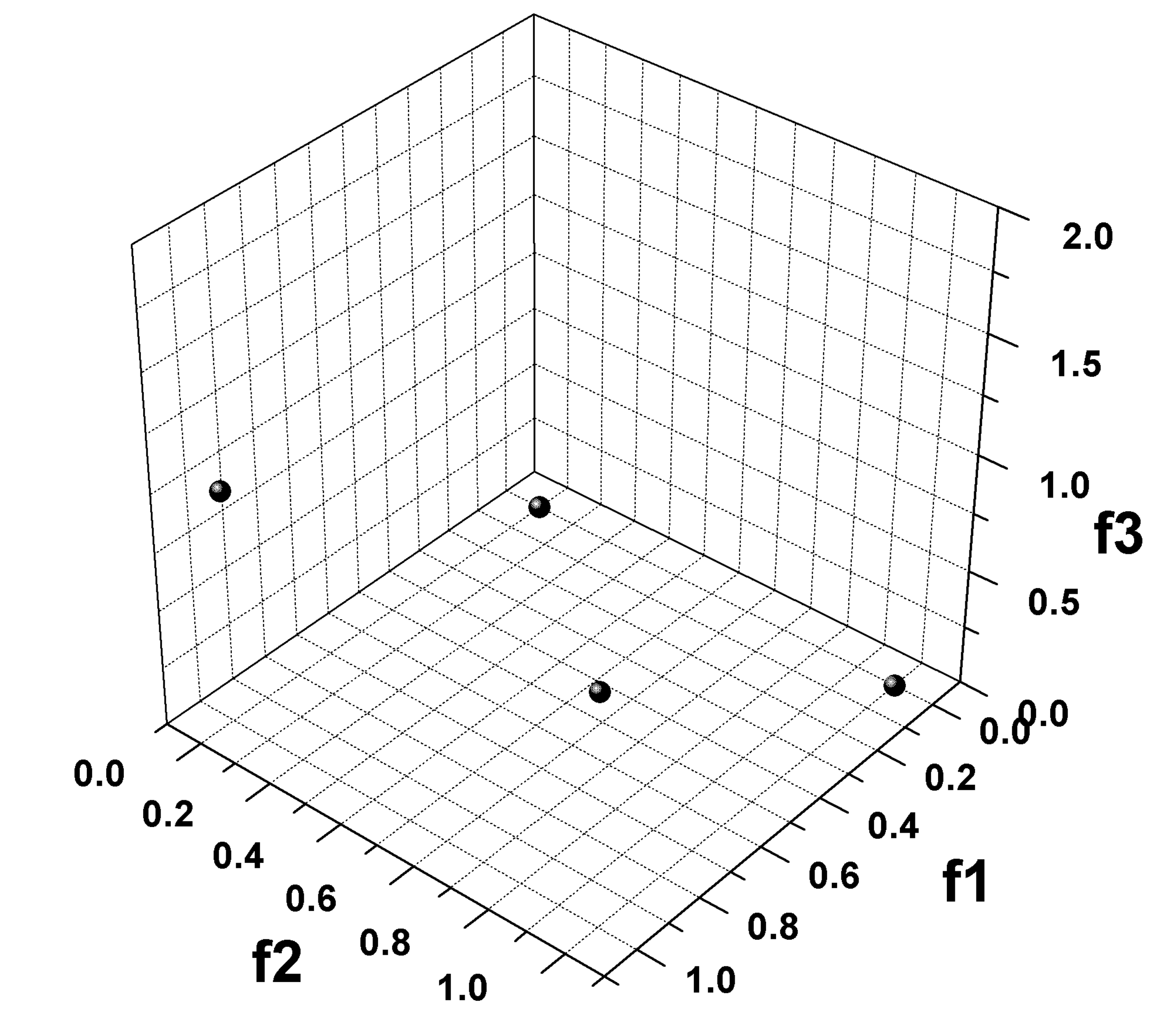} ~~~&~~~
			\includegraphics[scale=0.22]{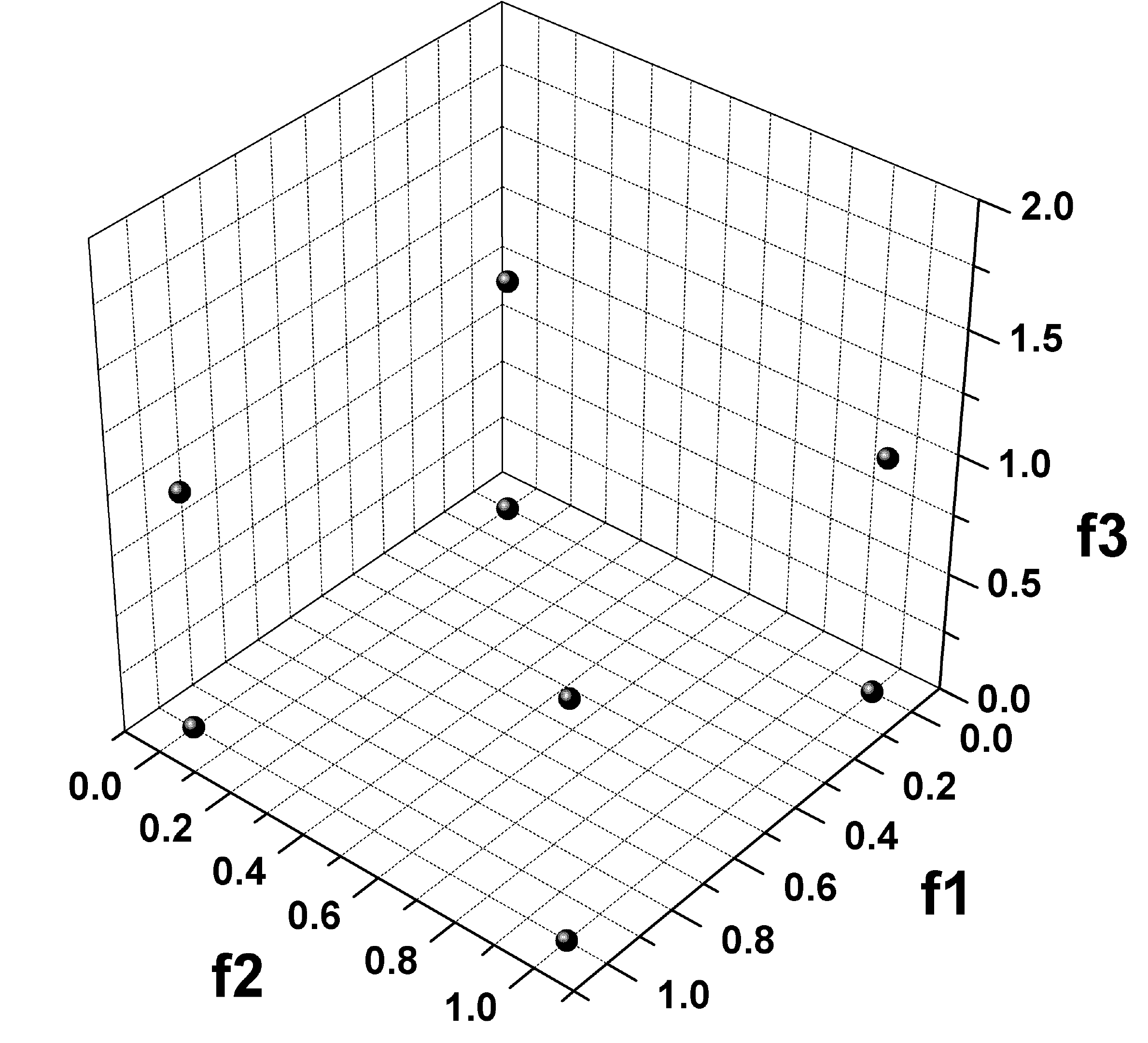}\\
			(a) Solution set A  & (b) Solution set B  \\
		\end{tabular}
	\end{center}
	\vspace{-3mm}
	\caption{An artificial example of two solution sets (A and B) having same parallel coordinates plots shown in \mbox{Figure~\ref{Fig:ExampleCoverage}}.}
	\label{Fig:ExampleCoverageCC}
\end{figure}
%%%% Fig. 8 %%%%
\begin{figure}[!]
	\begin{center}
		\footnotesize
		\begin{tabular}{@{}cc}
			\includegraphics[scale=0.22]{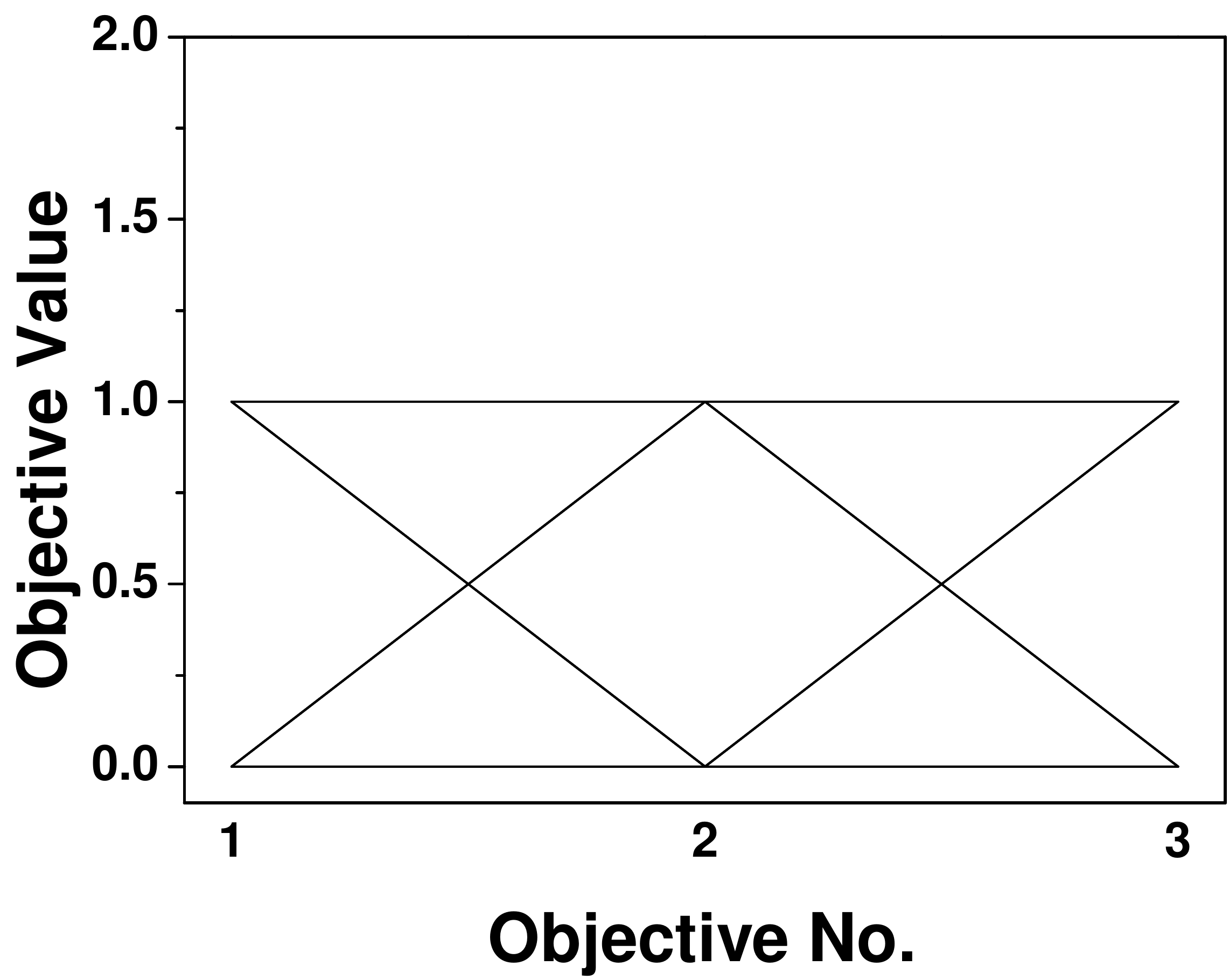} ~~~&~~~
			\includegraphics[scale=0.22]{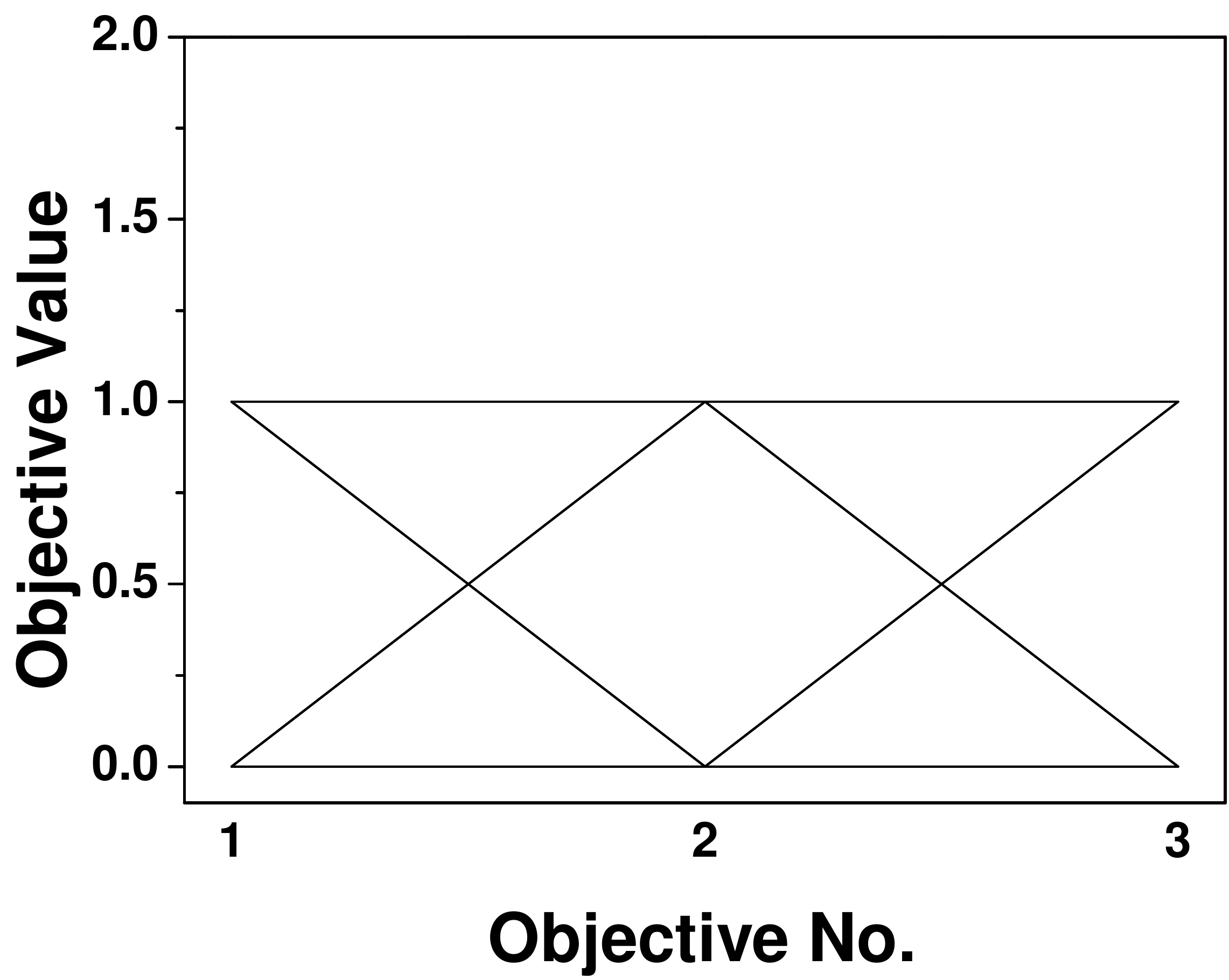}\\
			(a) Solution set A  & (b) Solution set B  \\
		\end{tabular}
	\end{center}
	\vspace{-3mm}
	\caption{The parallel coordinates of the solution sets in \mbox{Figure~\ref{Fig:ExampleCoverageCC}}.}
	\label{Fig:ExampleCoverage}
\end{figure}

%%%% Fig. 9 %%%%
\begin{figure}[tbp]
	\begin{center}
		\footnotesize
		\begin{tabular}{@{}cc}
			\includegraphics[scale=0.22]{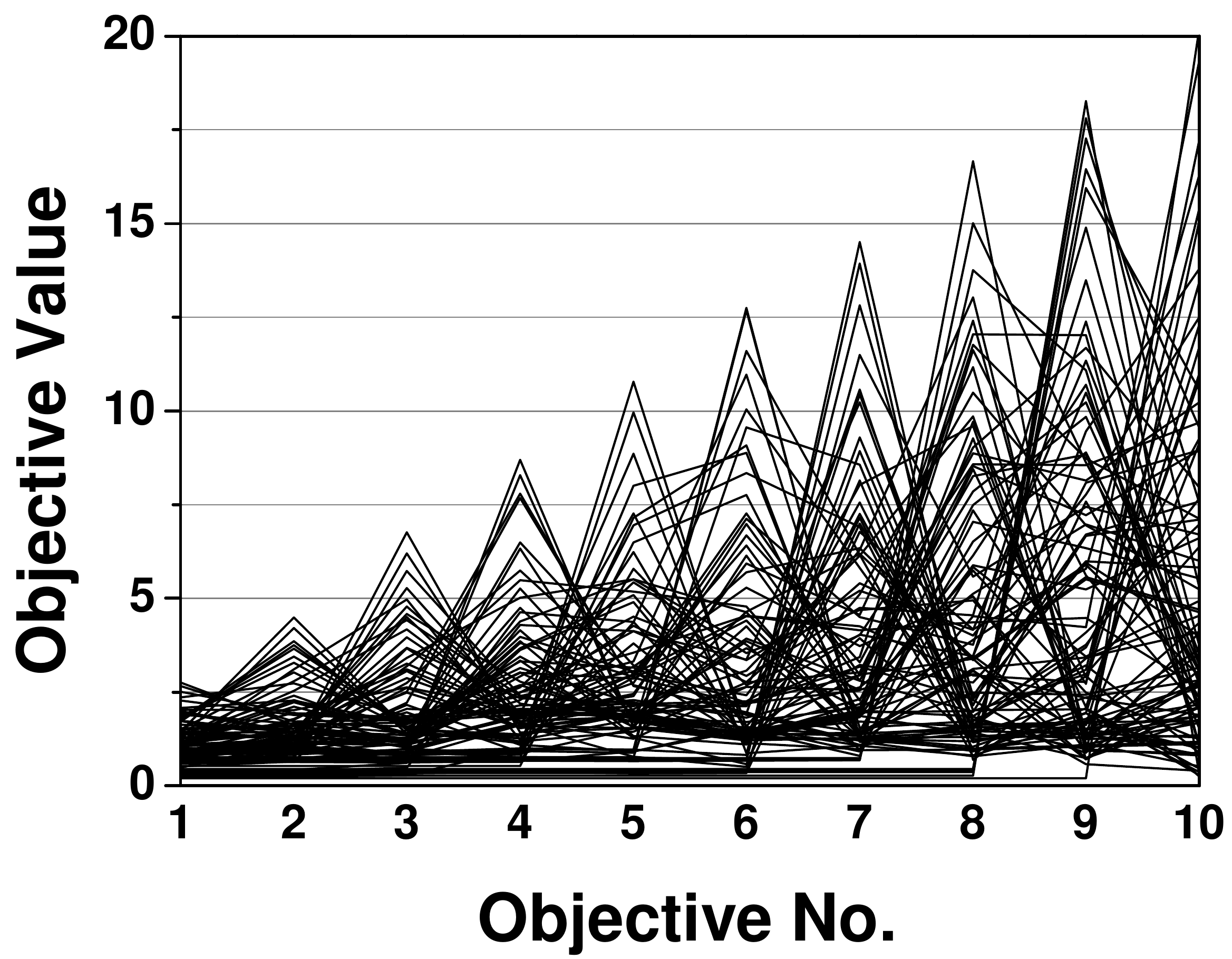} ~~~&~~~
			\includegraphics[scale=0.22]{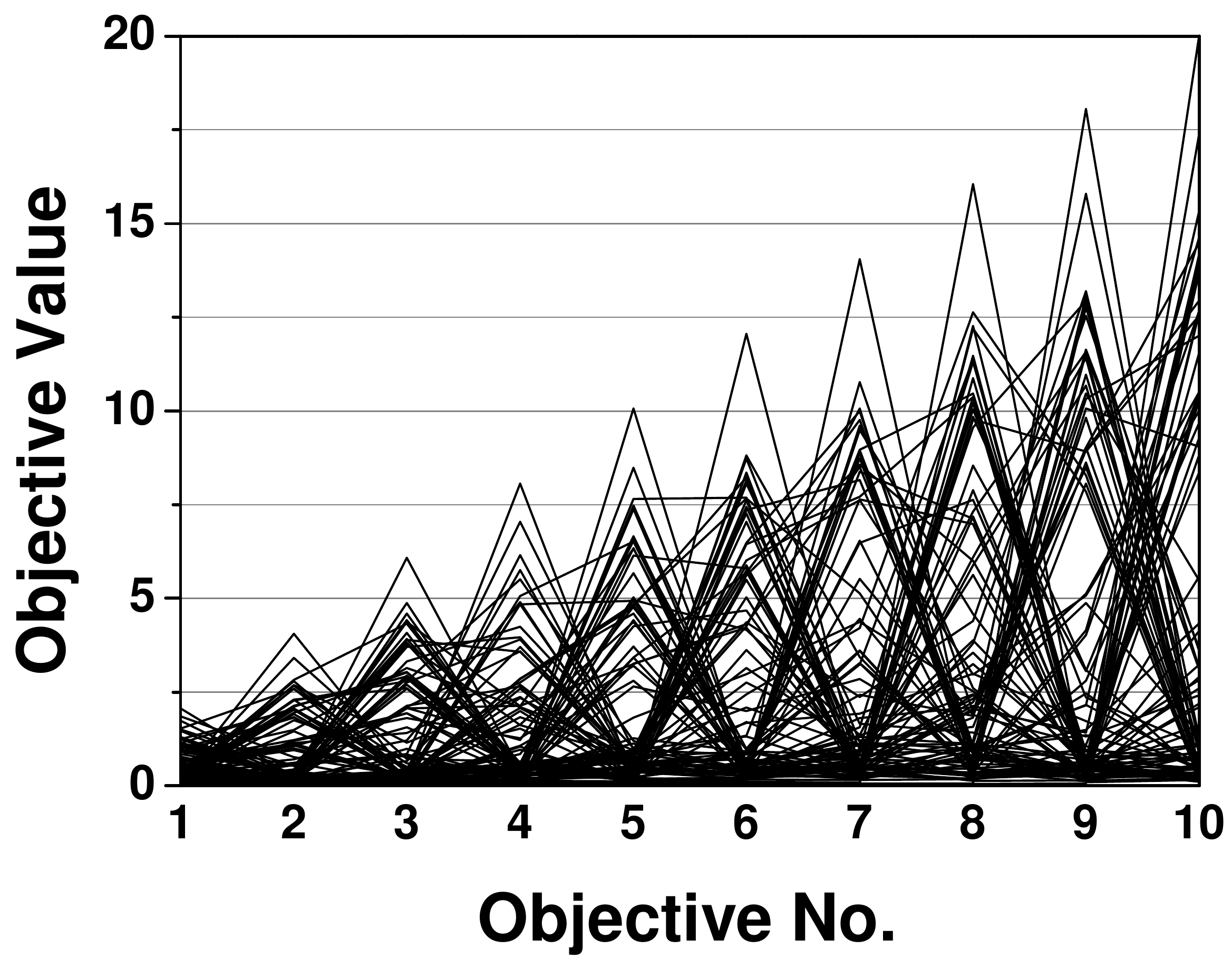}\\
			(a) NSGA-II (DCI=7.17E-1) & (b) GrEA (DCI=7.75E-1) \\
		\end{tabular}
	\end{center}
	\vspace{-3mm}
	\caption{The solution set obtained by NSGA-II and GrEA on the
		10-objective WFG7, 
		and their evaluation results on the coverage metric DCI (the bigger the better).}
	\label{Fig:WFG3coverage}
\end{figure}

In parallel coordinates, 
it is straightforward to see which region a solution set does not reach on any objective\footnote{Note that
for real-world problems whose Pareto front is unknown, 
we cannot tell if a solution set reaches the optimal region of objectives or not.}. 
For example, 
in \mbox{Figure~\ref{Fig:DTLZ210coverage}} the solution set obtained 
by the AR method~\cite{Bentley1997} concentrates in one tiny area 
and the set by IBEA~\cite{Zitzler2004} fails to cover the first six objectives on the 10-objective DTLZ2.
Moreover, 
we can conjecture some distribution features of solution sets from their parallel coordinates representation. 
Take the solution sets of \mbox{Figure~\ref{Fig:DTLZ2CoverageCC}} as an example;
their parallel coordinates representation is shown in \mbox{Figure~\ref{Fig:DTLZ2Coverage}}.
From \mbox{Figure~\ref{Fig:DTLZ2Coverage}}, 
we can know that the solution sets of IBEA and SMS-EMOA~\cite{Beume2007} fail to 
cover the region between 0 and 0.2 on all three objectives. 
Also, 
most of the solutions obtained by IBEA are located in the boundary of the Pareto front 
as there are very few lines distributed around the middle section on all three objectives in the figure. 

However, 
there do exist some cases that different solution sets have the same parallel coordinates plots.
We can easily construct such an example.
In \mbox{Figure~\ref{Fig:ExampleCoverageCC}}, 
solution set B has a better coverage than set A (the four solutions in set A being duplicate), 
but the two sets have the same pattern in parallel coordinates (\mbox{Figure~\ref{Fig:ExampleCoverage}}).
Note that if we change the order of some objectives (e.g., $f_1$ and $f_2$),
the parallel coordinates of the two solution sets in this example would be different.

One important fact that we would like to note is that 
as parallel coordinates map an $m$-dimensional graph onto a 2D graph 
they cannot fully reflect the coverage of solution sets.
A set of solutions (represented by polylines) may have a good coverage over 
the range of the Pareto front in the 2D graph, 
but they may only cover part of the Pareto front in the original $m$-dimensional space.
An interesting example is shown in \mbox{Figure~\ref{Fig:WFG3coverage}}.
In that figure,
NSGA-II appears to have a better coverage than GrEA according to the parallel coordinates plots,
but GrEA has a better coverage evaluation result,
measured by the coverage metric \textit{Diversity Comparison Indicator} (DCI)~\cite{Li2014d}.

\subsection{Uniformity}

\begin{figure}[tbp]
	\begin{center}
		\footnotesize
		\begin{tabular}{@{}c@{}c@{}}
			\hspace{-3mm}\includegraphics[scale=0.22]{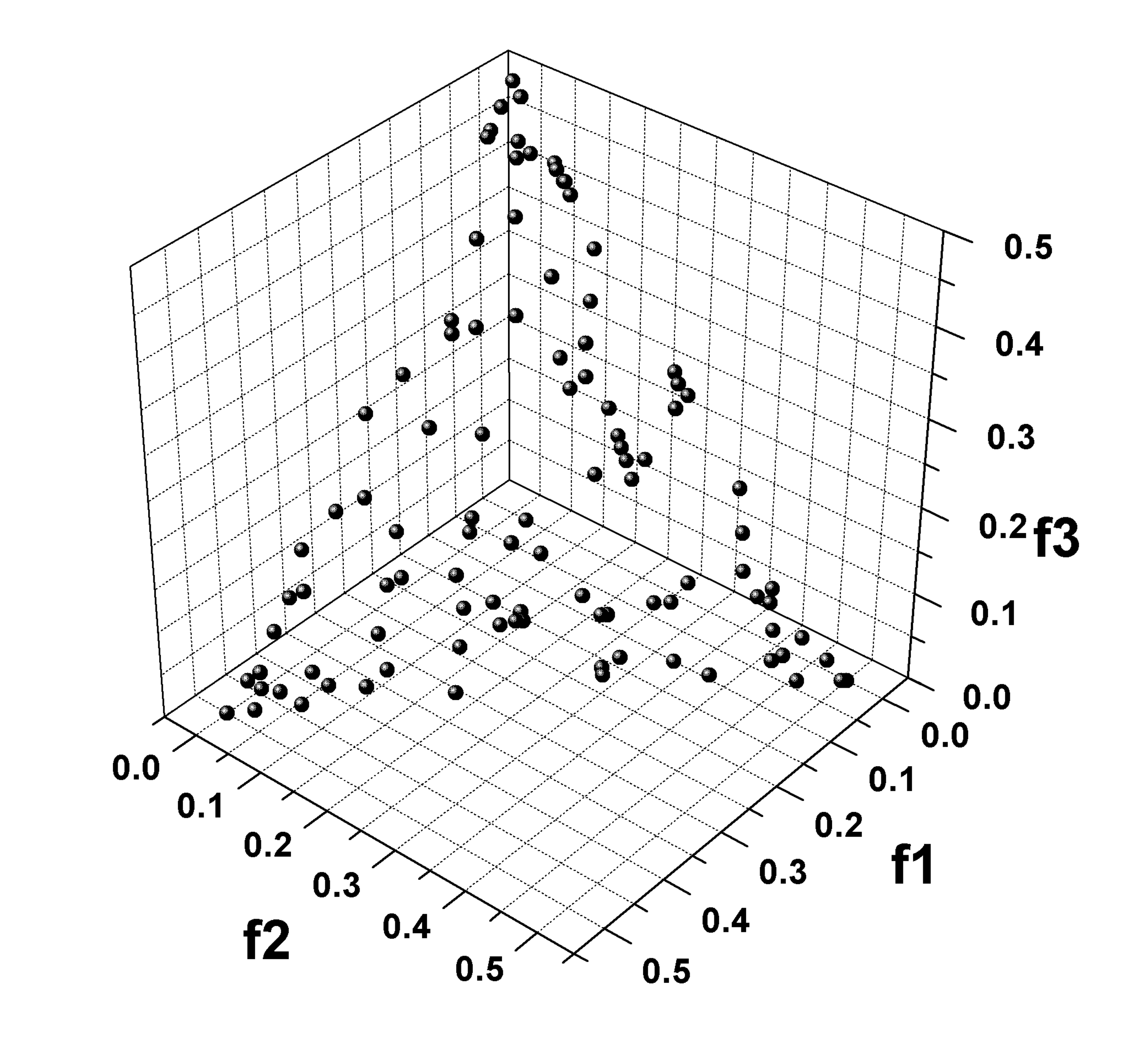} ~~~&~~~
			\includegraphics[scale=0.22]{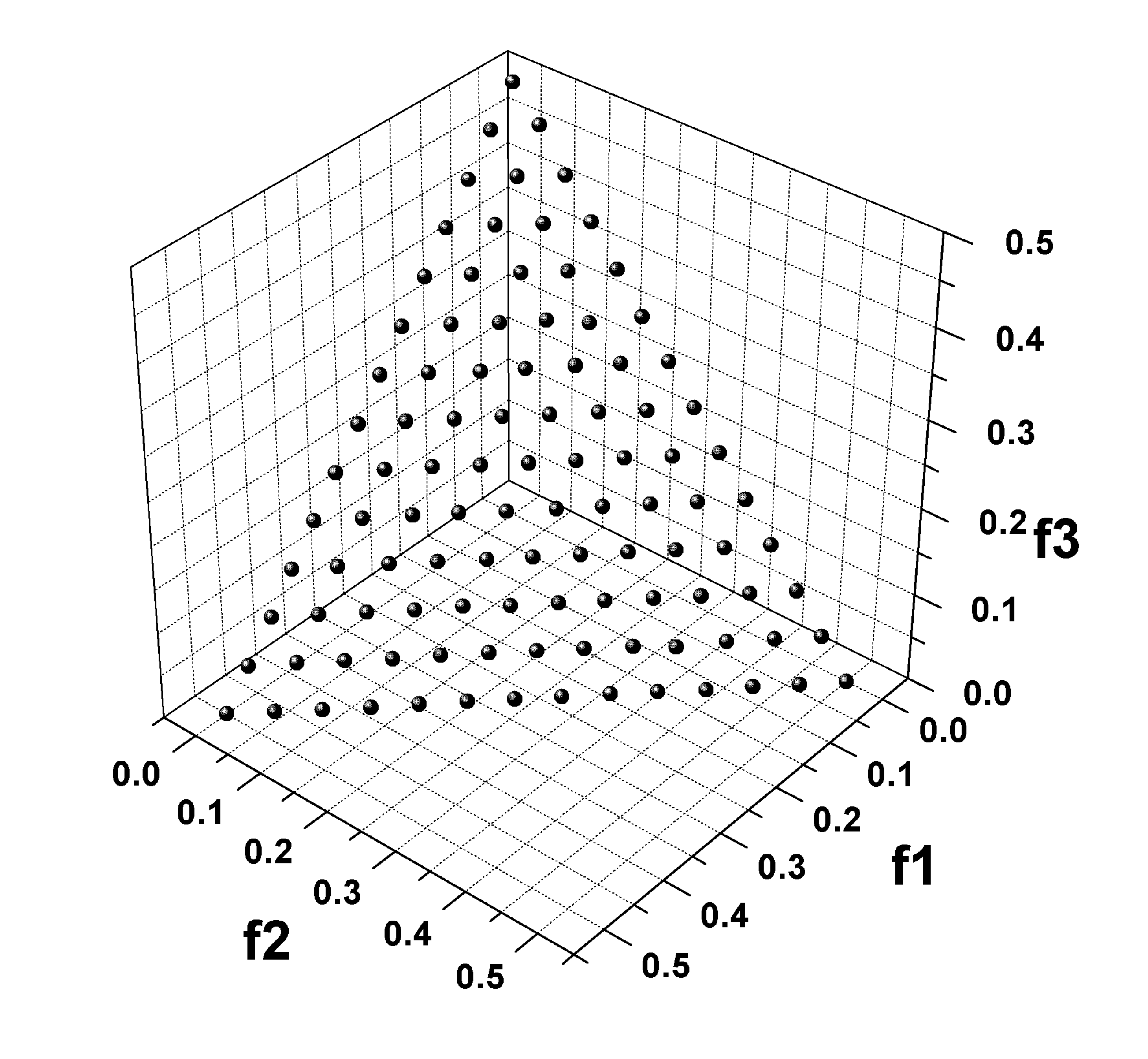}\\
			(a) NSGA-II & (b) MOEA/D \\
		\end{tabular}
	\end{center}
	\vspace{-3mm}
	\caption{The solution set obtained by NSGA-II and MOEA/D on the 3-objective DTLZ1, 
		shown in Cartesian coordinates.}
	\label{Fig:DTLZ1uniformityCC}
\end{figure}

\begin{figure}[tbp]
	\begin{center}
		\footnotesize
		\begin{tabular}{@{}cc}
			\includegraphics[scale=0.22]{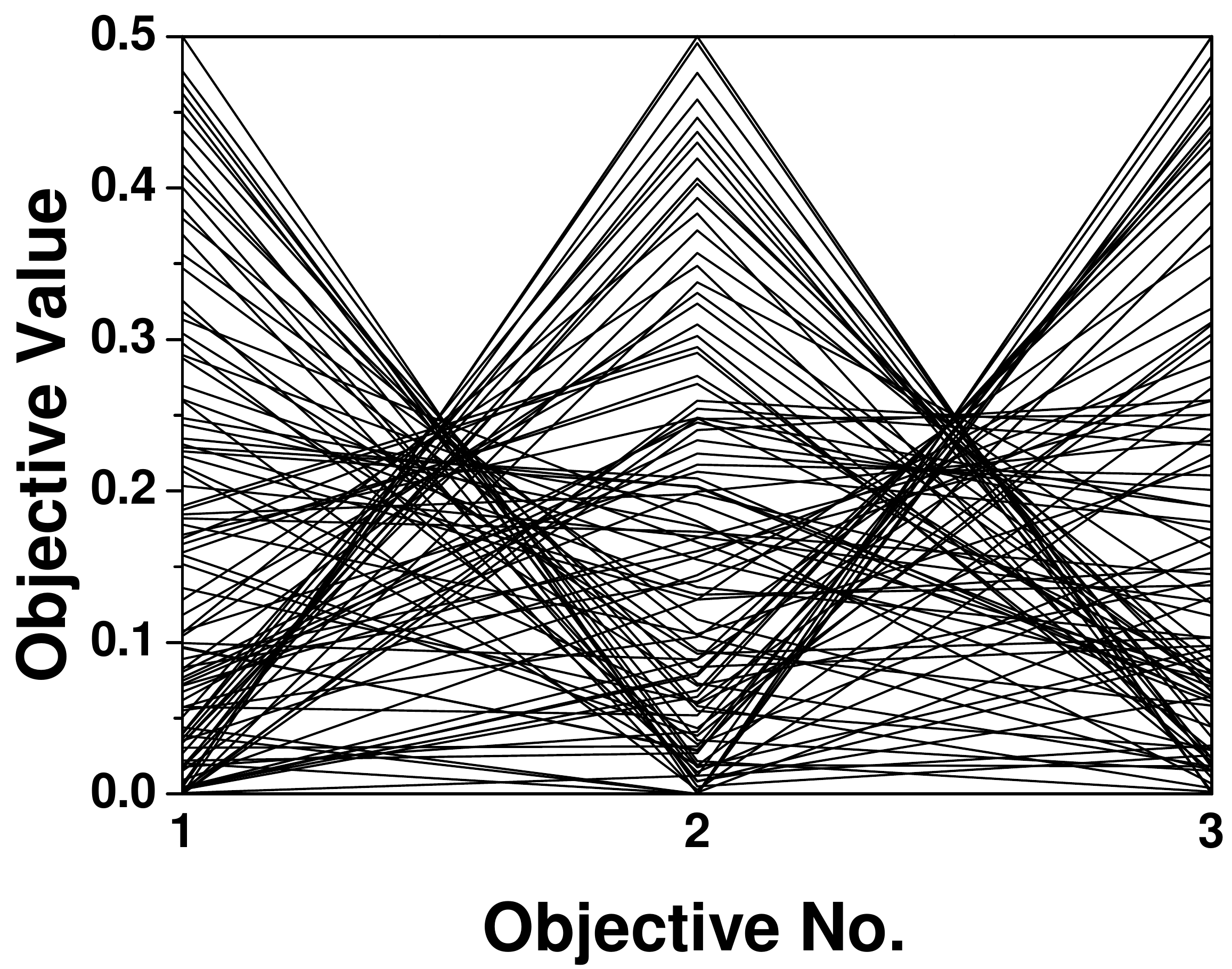}~~~&~~~
			\includegraphics[scale=0.22]{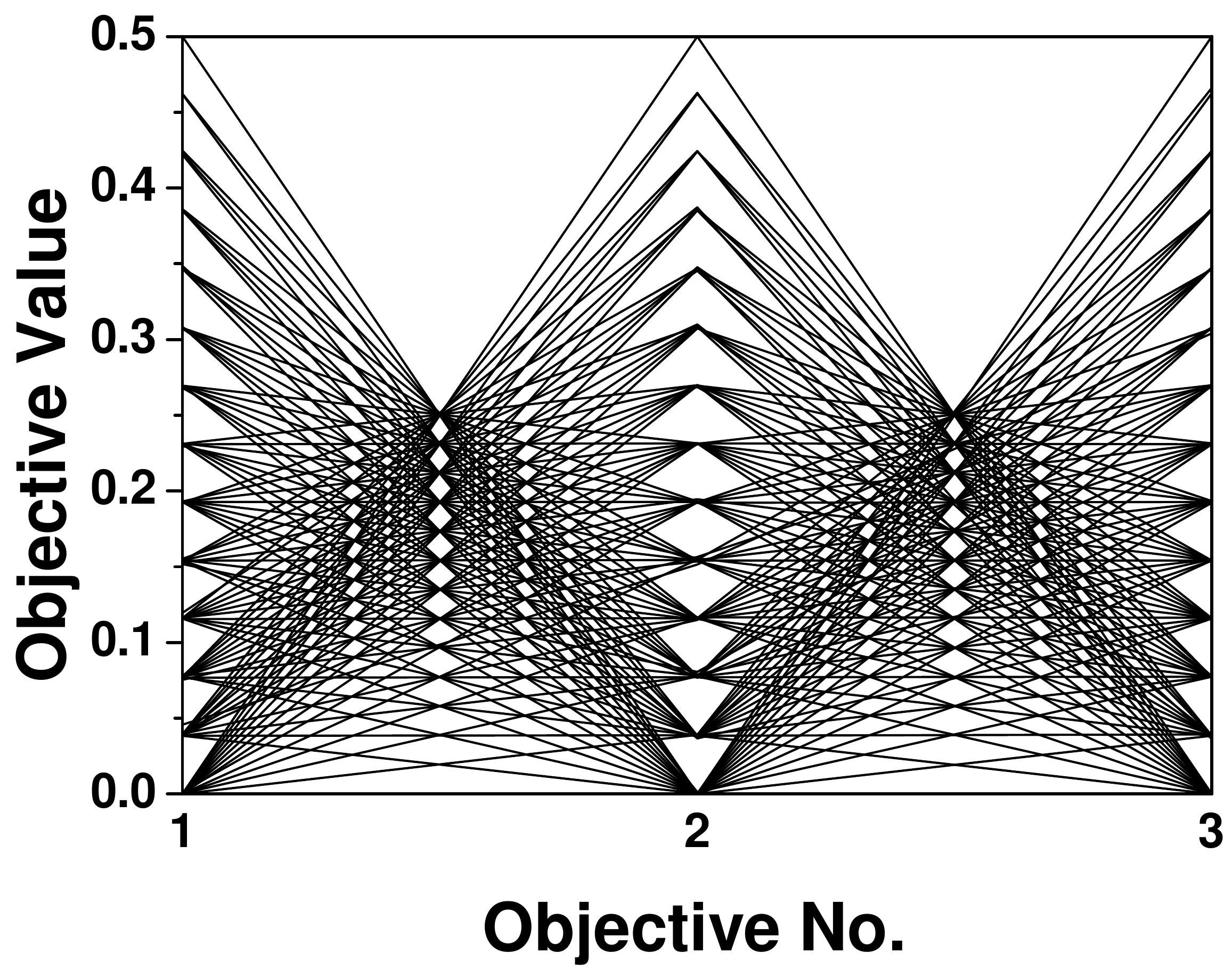}\\
			(a) NSGA-II & (b) MOEA/D \\
		\end{tabular}
	\end{center}
	\vspace{-3mm}
	\caption{The corresponding parallel coordinates of the solution sets in \mbox{Figure~\ref{Fig:DTLZ1uniformityCC}}.}
	\label{Fig:DTLZ1uniformity}
\end{figure}

\begin{figure}[!]
	\begin{center}
		\footnotesize
		\begin{tabular}{@{}cc}
			\includegraphics[scale=0.22]{DTLZ2MOEADCC.pdf}~~~&~~~
			\includegraphics[scale=0.22]{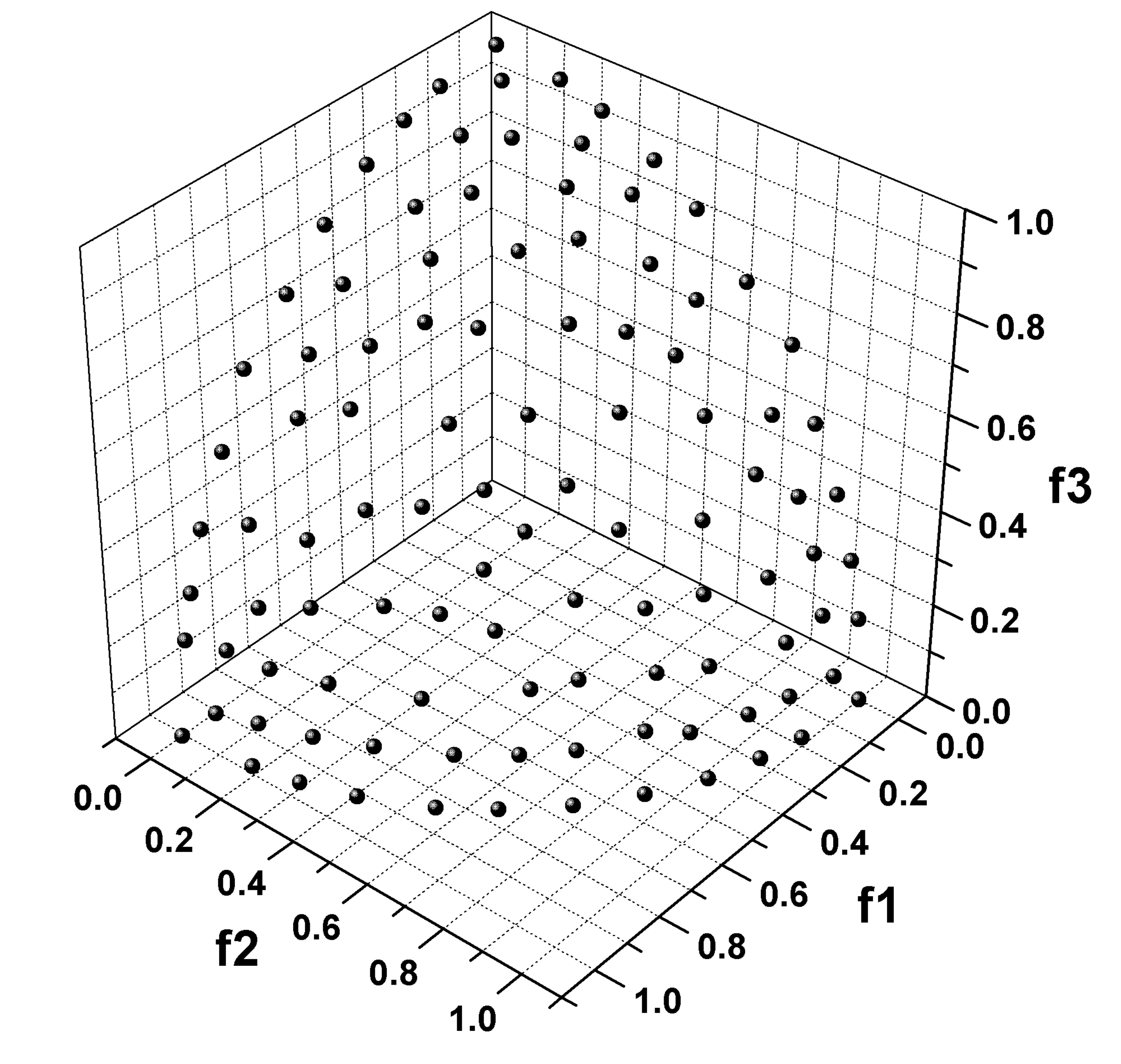}\\
			(a) MOEA/D & (b) BCE-MOEA/D \\
		\end{tabular}
	\end{center}
	\vspace{-3mm}
	\caption{The solution set obtained by MOEA/D and BCE-MOEA/D on the 3-objective DTLZ2, 
		shown in Cartesian coordinates.}
	\label{Fig:DTLZ2uniformityCC}
\end{figure}

\begin{figure}[tb]
	\begin{center}
		\scriptsize
		\begin{tabular}{@{}cc}
			\includegraphics[scale=0.22]{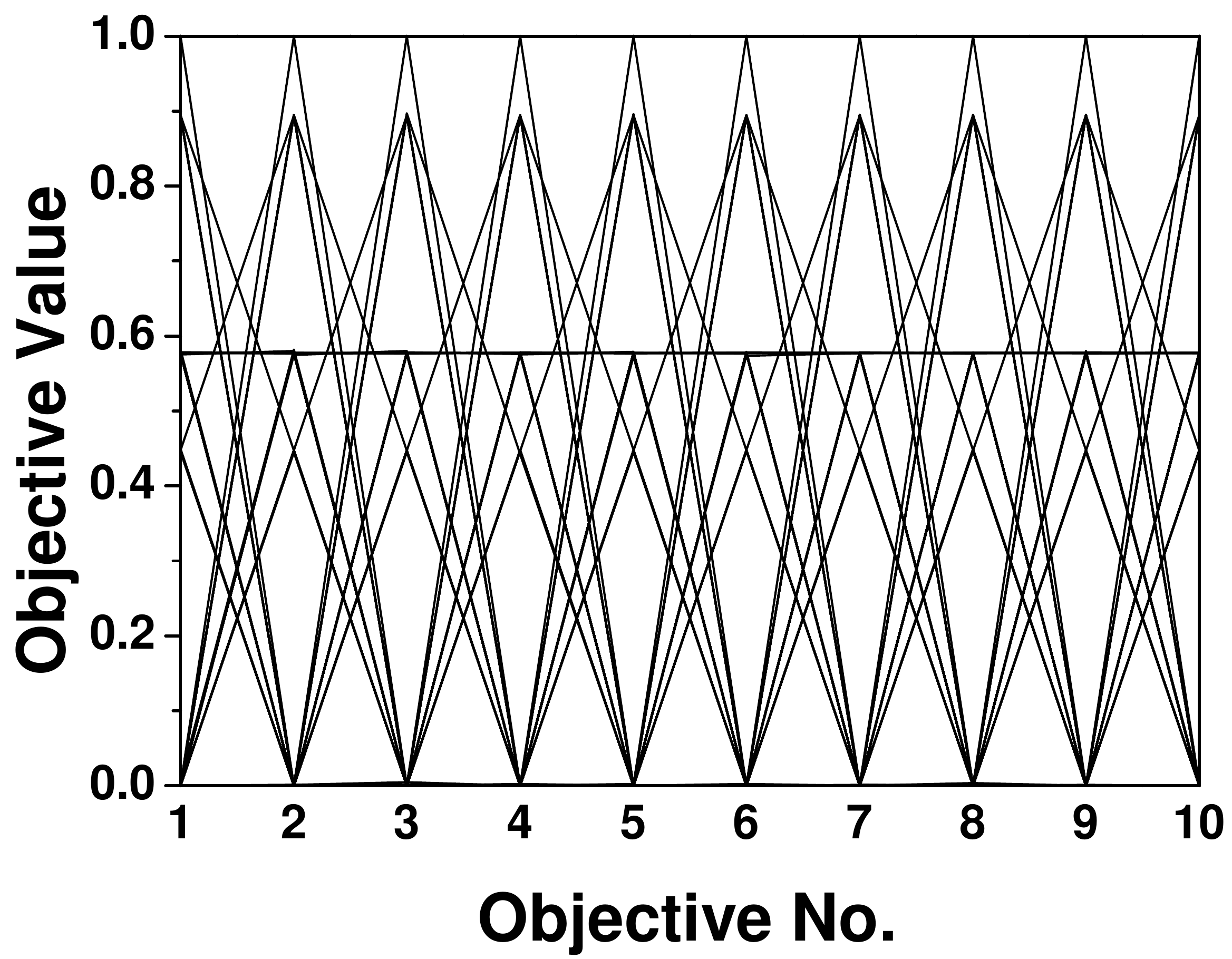}~~~&~~~
			\includegraphics[scale=0.22]{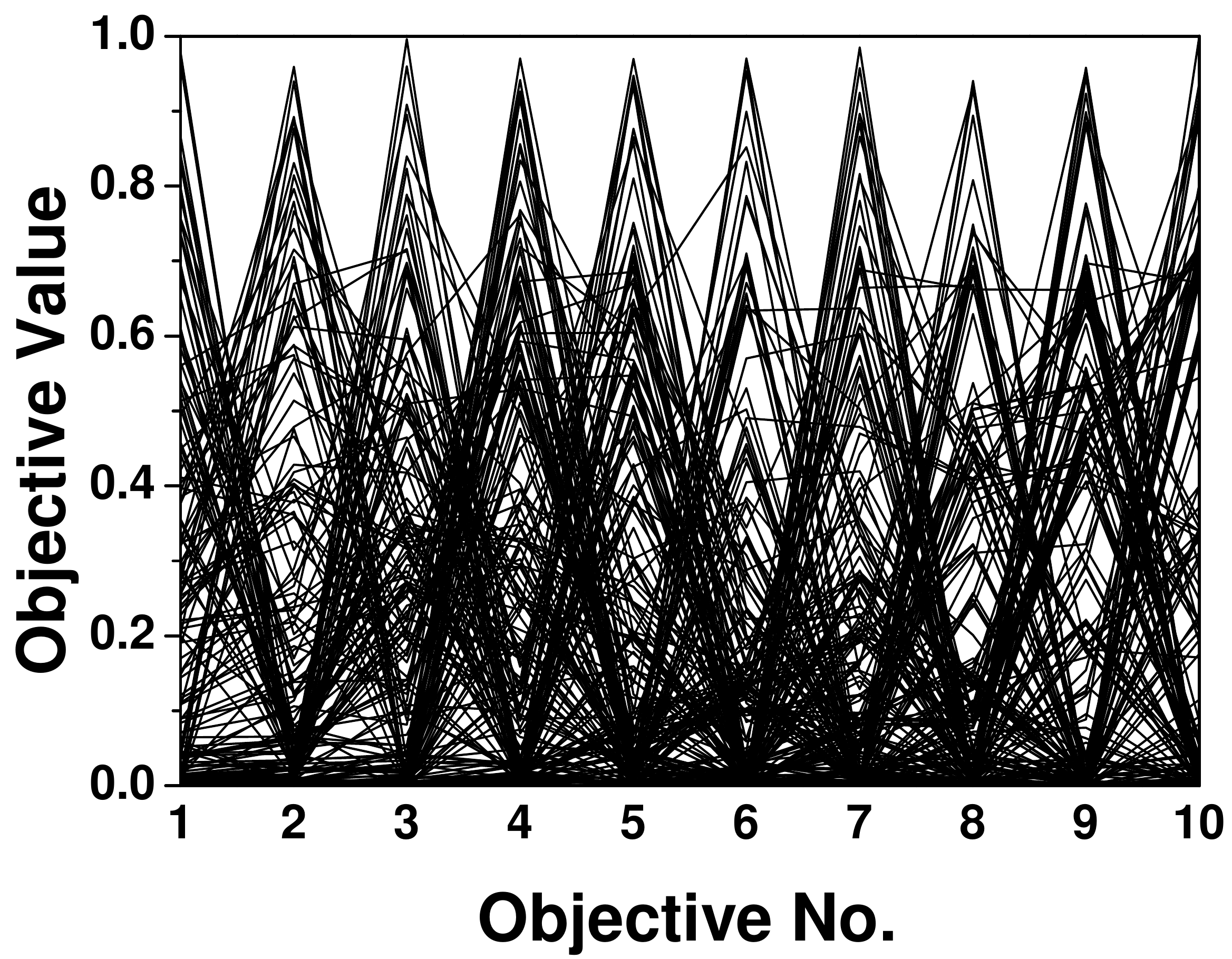}\\
			(a) MOEA/D (SP=1.05E-1) & (b) BCE-MOEA/D (SP=7.74E-2) \\
		\end{tabular}
	\end{center}
	\vspace{-3mm}
	\caption{The solution set obtained by MOEA/D and BCE-MOEA/D on the 3-objective DTLZ2, 
		and their evaluation results on the uniformity metric SP (the smaller the better).}
	\label{Fig:DTLZ2uniformity}
\end{figure}

In parallel coordinates, 
it is not easy to see how evenly a set of solutions are distributed. 
However, 
a set of uniformly-distributed polylines in parallel coordinates often implies a uniformly-distributed solution set.
As shown in \mbox{Figures~\ref{Fig:DTLZ1uniformityCC} and \ref{Fig:DTLZ1uniformity}}, 
MOEA/D~\cite{Zhang2007} has a perfectly-distributed solution set and 
its corresponding polylines in parallel coordinates are distributed uniformly and regularly.
This is in contrast to the solution set of NSGA-II 
which are distributed rather irregularly in both Cartesian and parallel coordinates plots.
Note that a set of irregularly-distributed polylines may not represent a badly-distributed solution set, 
as uniformly-distributed solutions can have distinct values on different objectives.
To show this, 
we select two EMO algorithms, 
MOEA/D~\cite{Zhang2007} and BCE-MOEA/D~\cite{Li2016},
both of which are able to obtain a uniformly-distributed solution set on DTLZ2 (see \mbox{Figure~\ref{Fig:DTLZ2uniformityCC}}). 
In MOEA/D, 
the population distribution is maintained by a set of systematically-generated, 
uniformly-distributed weight vectors (within a simplex), 
and thus ideally its solutions only take several equivalent values on all the objectives. 
In contrast, 
in BCE-MOEA/D the population distribution is maintained by a niching-based criterion, 
and thus its solutions can spread over the whole range for each objective.
\mbox{Figure~\ref{Fig:DTLZ2uniformity}} gives the solution sets obtained by
MOEA/D and BCE-MOEA/D on the 10-objective DTLZ2\footnote{Here the number of DTLZ2's decision variables 
	is set to $m-1$ ($m$ is the number of objectives) 
	to ensure that all solutions produced by algorithms are Pareto optimal; 
	thus the uniformity measure cannot be affected by the difference of solution sets' convergence.}. 
As seen, 
on the uniformity metric \textit{Spacing} (SP)\footnote{In this paper, 
the SP metric has been slightly modified to make it compatible with Pareto dominance. 
That is, if two solution sets are comparable in terms of Pareto dominance, 
then the SP value of the dominating set is $0$ and the SP value of the dominated set is $1$.}\cite{Schott1995}, 
BCE-MOEA/D even performs better than MOEA/D, 
but we cannot see this from their parallel coordinates representation in the figure.
This phenomenon may happen frequently 
when comparing decomposition-based algorithms having a set of systematically-generated weight vectors 
(such as MOEA/D and NSGA-III~\cite{Deb2014})
with algorithms that do not use such decomposition techniques (such as SPEA2~\cite{Zitzler2002} and Two\_arch2~\cite{Wang2015}).
So care needs to be taken when making a conclusion 
about the distribution uniformity of solution sets from parallel coordinates.

%%%% Fig. 10 %%%%

%%%% Fig. 11 %%%%
%%%% Fig. 12 %%%%

\begin{figure}[tbp]
	\begin{center}
		\footnotesize
		\begin{tabular}{@{}cc}
			\includegraphics[scale=0.22]{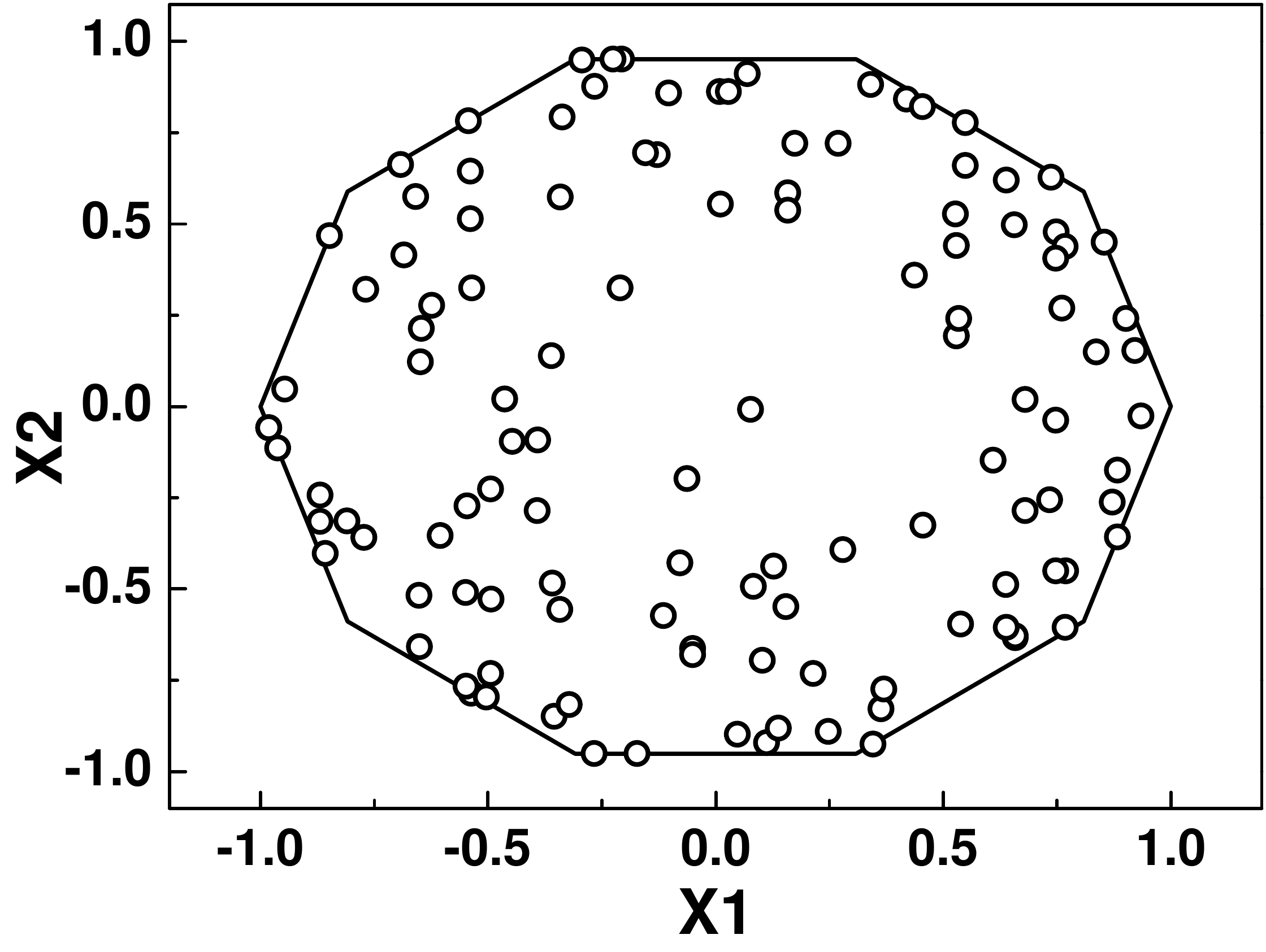}~~~&~~~
			\includegraphics[scale=0.22]{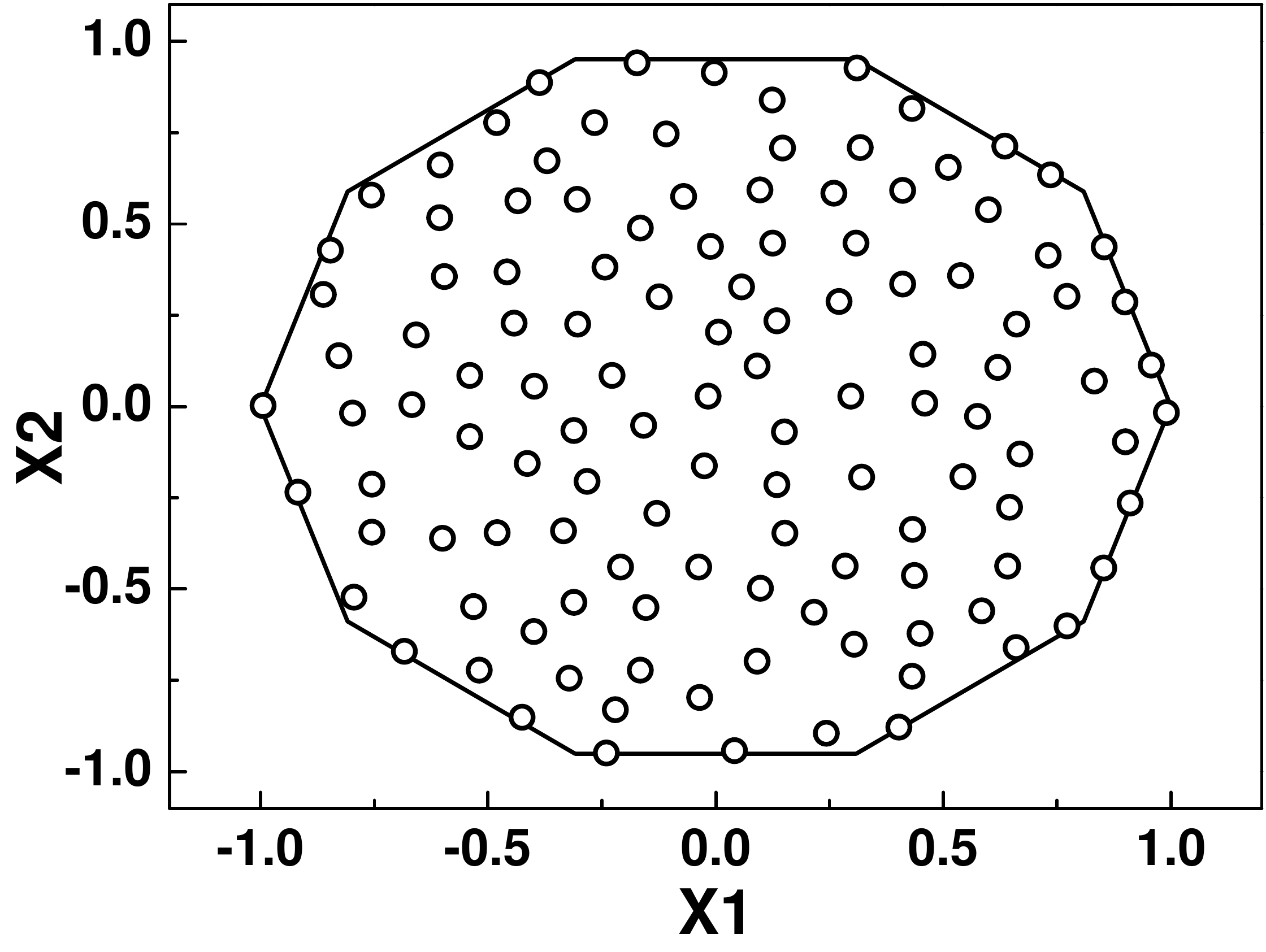}\\
			(a) NSGA-II & (b) SPEA2 \\
		\end{tabular}
	\end{center}
	\vspace{-3mm}
	\caption{The solution sets (in the decision space) obtained by NSGA-II and SPEA2 on the
		10-objective ML-DMP where the search space is precisely the optimal polygon.}
	\label{Fig:MLDMPvar}
\end{figure}

\begin{figure}[tbp]
	\begin{center}
		\footnotesize
		\begin{tabular}{@{}cc}
			\includegraphics[scale=0.22]{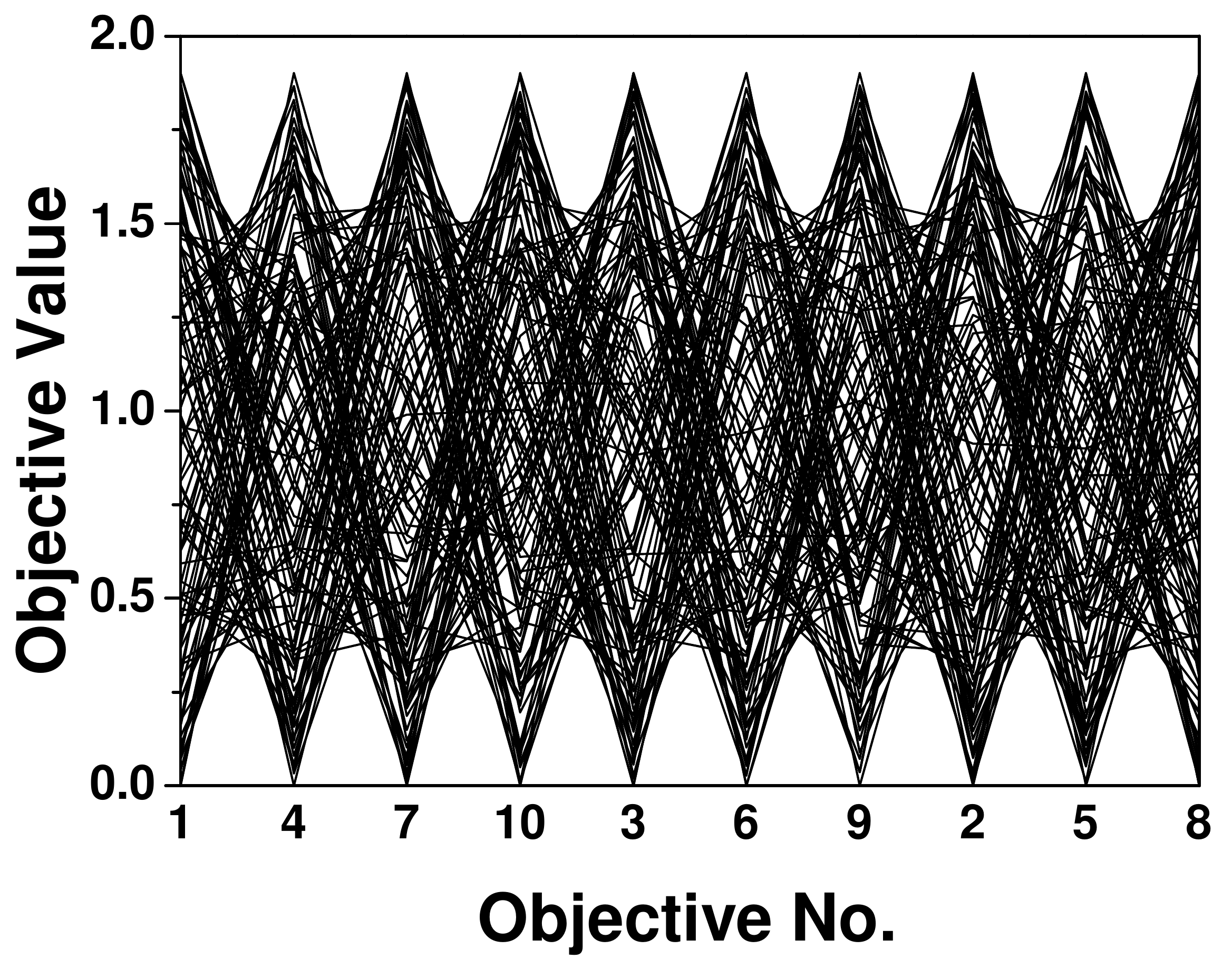}~~~&~~~
			\includegraphics[scale=0.22]{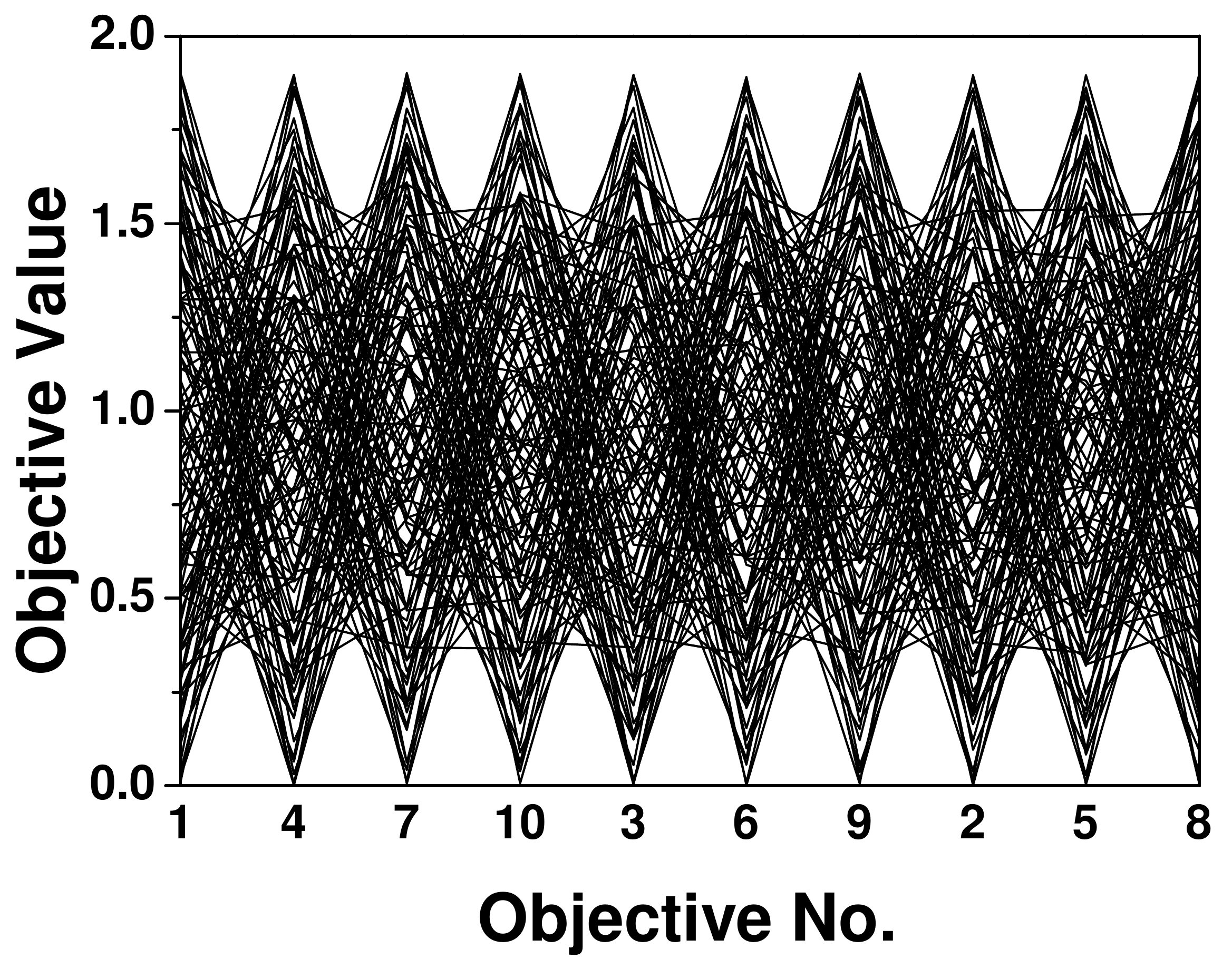}\\
			(a) NSGA-II & (b) SPEA2 \\
		\end{tabular}
	\end{center}
	\vspace{-3mm}
	\caption{Parallel coordinates of the solution sets (in the objective space) in \mbox{Figure~\ref{Fig:MLDMPvar}}.}
	\label{Fig:MLDMP}
\end{figure}

Finally,
it is worth mentioning that parallel coordinates plots can be easily cluttered with multiple lines overlaid.
This may completely prevent solution sets' distribution from being observed.
\mbox{Figures~\ref{Fig:MLDMPvar} and \ref{Fig:MLDMP}} show such an example, 
with two solution sets obtained by NSGA-II and SPEA2 on the 10-objective ML-DMP problem~\cite{Li2014c,Li2017}.  
The $m$-objective ML-DMP minimizes the distance of two-dimensional points to a set of $m$ straight lines, 
each of which passes through one edge of a given regular polygon with $m$ vertexes.
One interesting characteristic of ML-DMP is that the points in the regular polygon and 
their objective images are similar in the sense of Euclidean geometry. 
In other words, 
the ratio of the distance between any two points in the polygon to the distance
between their corresponding objective vectors is a constant. 
This allows a straightforward understanding of the distribution of the objective vector set 
via observing the solution set in the 2D decision space.
As can be seen in \mbox{Figure~\ref{Fig:MLDMPvar}},
SPEA2 has a far better distribution uniformity than NSGA-II,
but we cannot see the difference between their parallel coordinates representation in \mbox{Figure~\ref{Fig:MLDMP}}.

%%%% Fig. 13 %%%%

\section{Solution Set Distributions}

%%%% Fig. 14 %%%%
%%%% Fig. 15 %%%%

\begin{figure}[tbp]
	\begin{center}
		\includegraphics[scale=0.25]{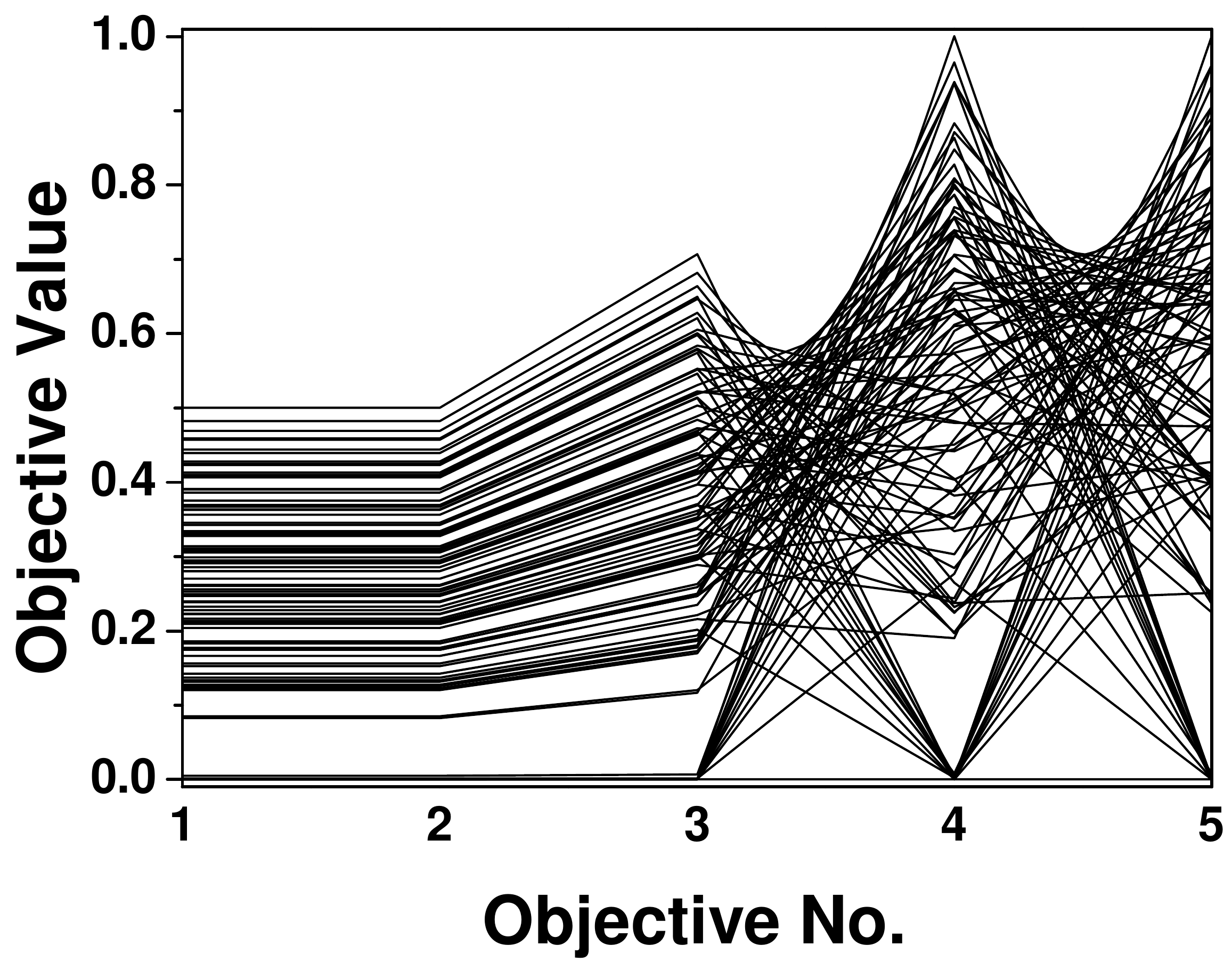}~~~~~
	\end{center}
	\vspace{-3mm}
	\caption{The solution set obtained by SPEA2+SDE~\cite{Li2014a} on the 5-objective DTLZ5($I,M$)~\cite{Deb2006}, 
		where $I=3$.}
	\label{Fig:DTLZ5IMshape}
\end{figure}

In parallel coordinates, 
it is straightforward to know the conflict between objectives.
The number of intersection lines between adjacent objectives reflects their conflicting degree.
If there is no intersection of any pair of lines between adjacent objectives, 
then these two objectives are completely non-conflicting (i.e., harmonious~\cite{Purshouse2003b}),
such as objectives $f_1$ versus $f_2$ and objectives $f_2$ versus $f_3$ 
in \mbox{Figure~\ref{Fig:DTLZ5IMshape}}.
If there are many lines intersecting,
then the two objectives are heavily conflicting, 
such as objectives $f_3$ versus $f_4$ and objectives $f_4$ versus $f_5$ in \mbox{Figure~\ref{Fig:DTLZ5IMshape}}.
If any pair of lines intersect, 
then the two objectives are completely conflicting to each other.

\begin{figure}[tbp]
	\begin{center}
		\footnotesize
		\begin{tabular}{@{}cc}
			\includegraphics[scale=0.23]{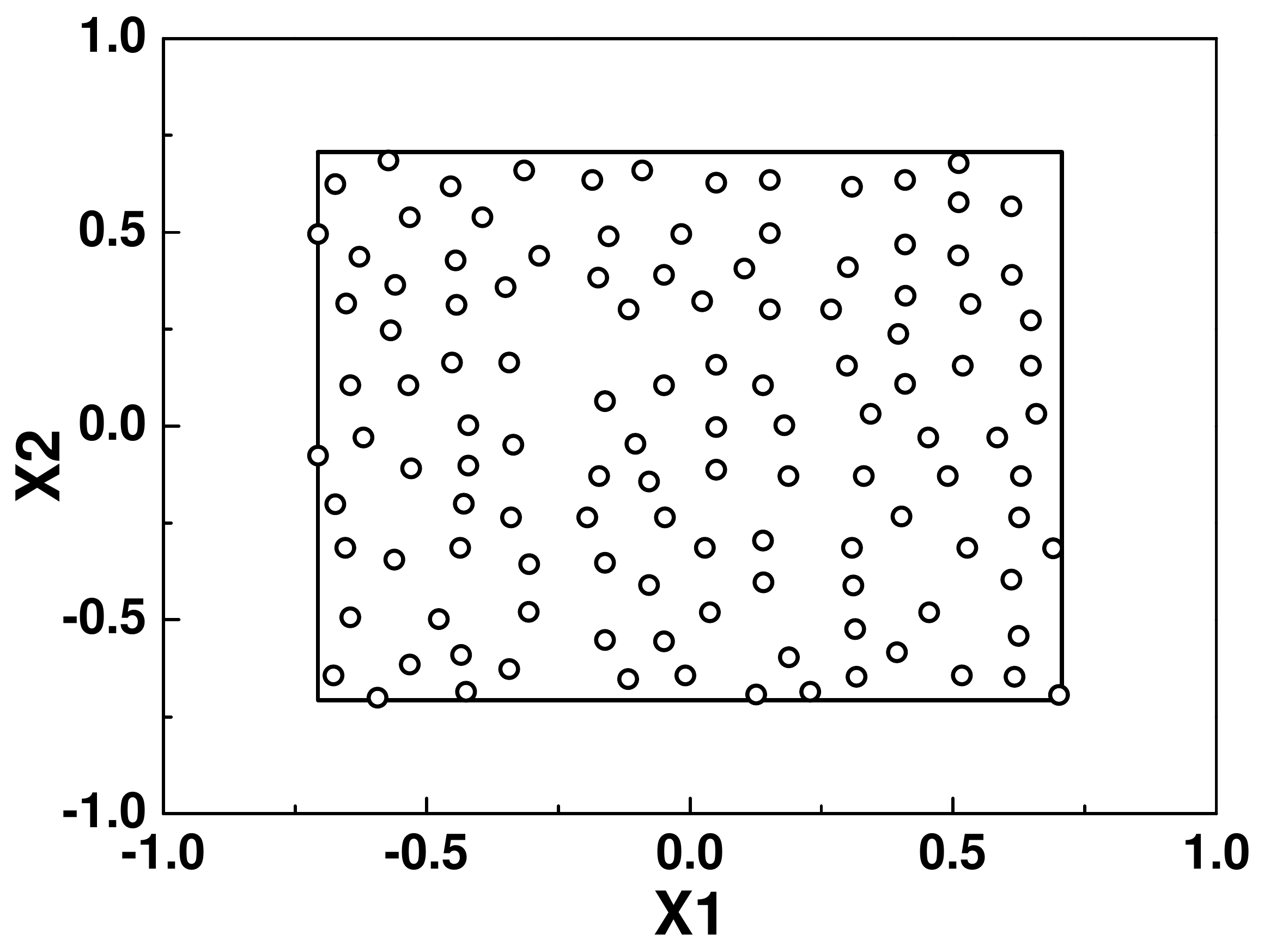}~~~&~~~
			\includegraphics[scale=0.225]{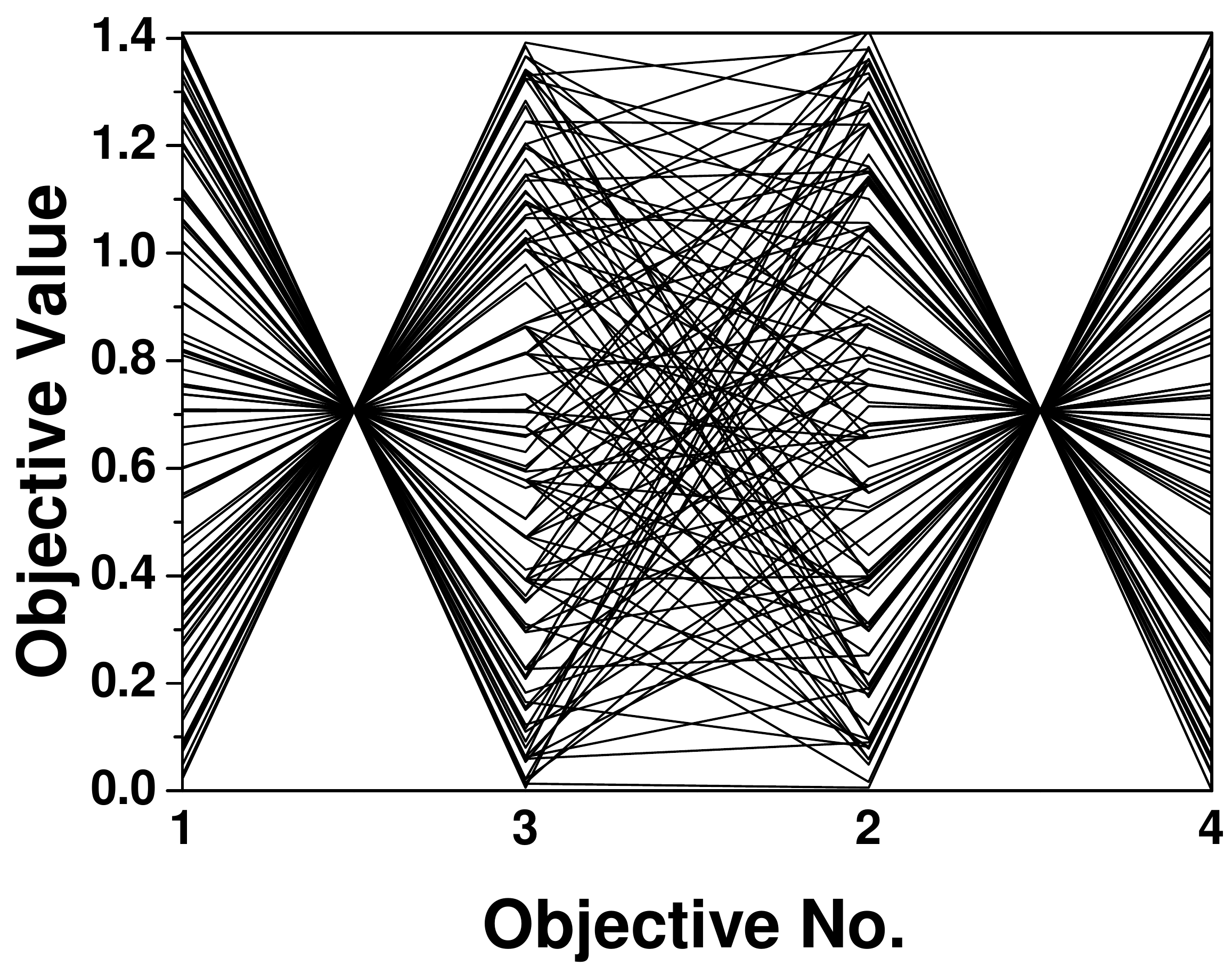}\\
			(a) Decision space & (b) Objective space \\
		\end{tabular}
	\end{center}
	\vspace{-3mm}
	\caption{The solution set of SPEA2+SDE on the 4-objective ML-DMP.}
	\label{Fig:MLDMPshapeSDE}
\end{figure}

An interesting phenomenon in the parallel coordinates plot is that 
if all lines between two adjacent objectives intersect at one point, 
then these two objectives are negatively linearly dependent.
\mbox{Figure~\ref{Fig:MLDMPshapeSDE}} is such an example. 
The four-objective ML-DMP problem minimizes the distance of points to four lines 
passing through the four edges of the given rectangle. 
From this definition, 
we can see that the two pairs of objectives, 
$f_1$ versus $f_3$ and $f_2$ versus $f_4$, 
are negatively linearly dependent for the solutions in the rectangle  
($f_1 = -f_3, f_2 = -f_4$).
Therefore, 
each of the objective pairs intersects at one point, 
as shown in \mbox{Figure~\ref{Fig:MLDMPshapeSDE}(b)}.

\begin{figure}[!]
	\begin{center}
		\footnotesize
		\begin{tabular}{@{}cc}
			\includegraphics[scale=0.48]{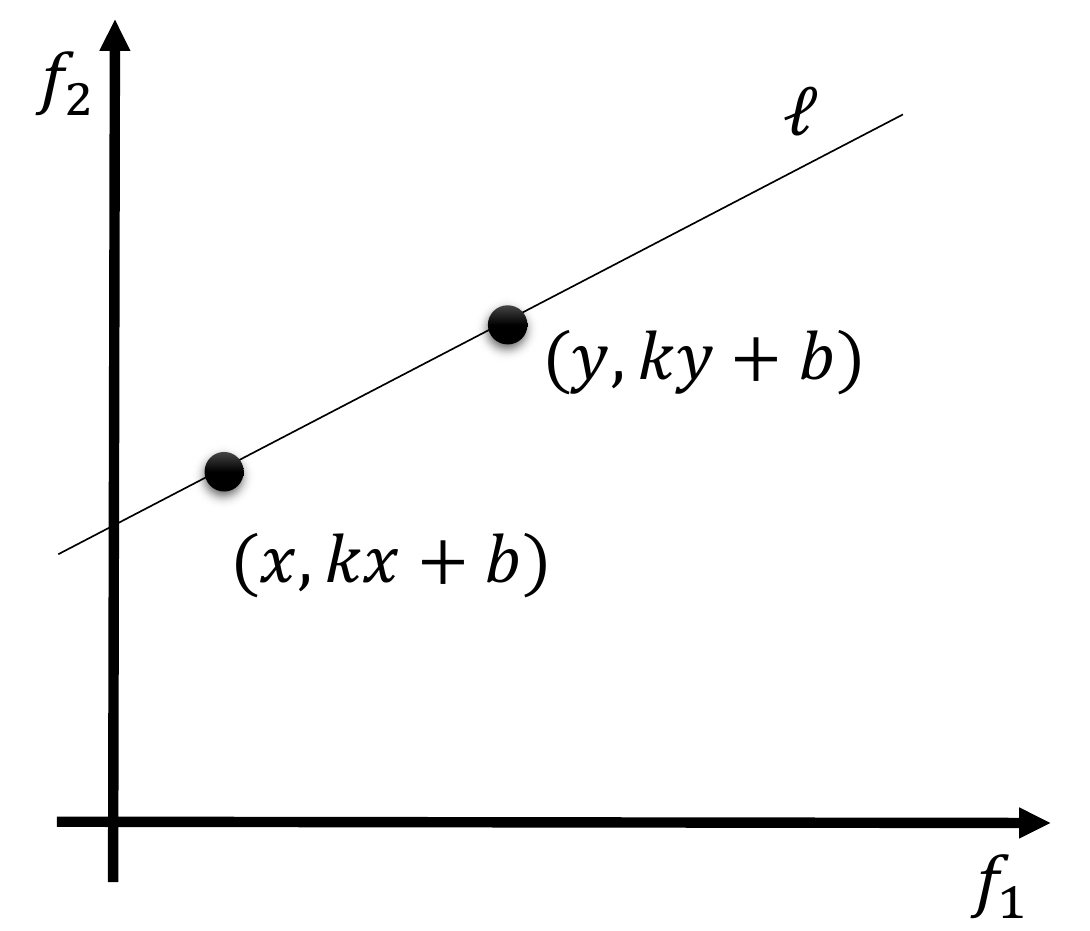}~~~&~~~
			\includegraphics[scale=0.48]{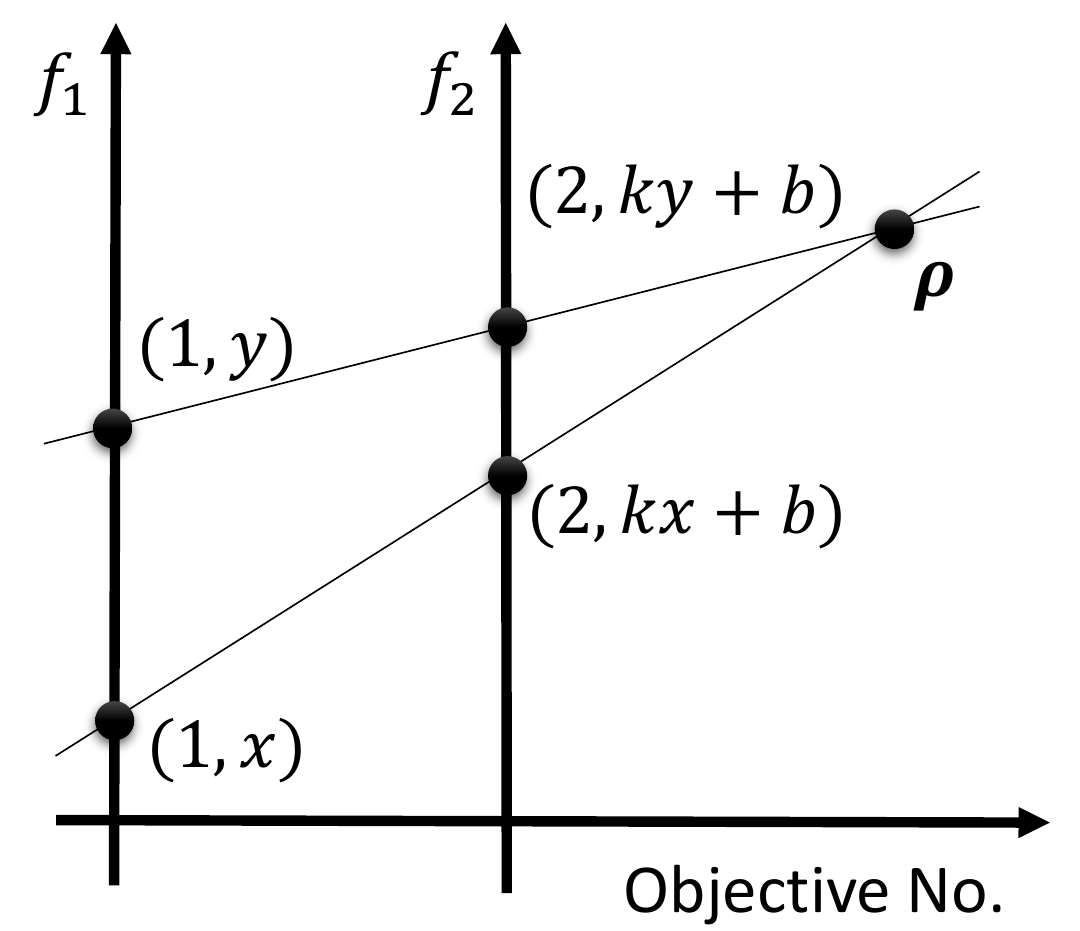}\\
			(a) Cartesian coordinates & (b) Parallel coordinates \\
		\end{tabular}
	\end{center}
	\vspace{-3mm}
	\caption{An example of a line $\ell: f_2 = kf_1+b$ in the Cartesian coordinate plane corresponding to a point 
		$\rho:((1-k)^{-1}, b(1-k)^{-1})$ in the parallel coordinates plane.} 
	\label{Fig:Cartesian2parallel}
\end{figure}

This property is the known duality between the parallel coordinates representation and 
the Cartesian coordinate representation of data~\cite{Inselberg1985,Wegman1990}: 
points in Cartesian coordinates map into lines in parallel coordinates,
while lines in Cartesian coordinates map into points in parallel coordinates.
Consider a line $\ell: f_2 = kf_1+b$ in the Cartesian coordinate plane of $f_1$ and $f_2$,
and consider two points lying on this line, 
say ($x, kx+b$) and ($y, ky+b$) (shown in \mbox{Figure~\ref{Fig:Cartesian2parallel}}(a)).
\mbox{Figure~\ref{Fig:Cartesian2parallel}(b)} shows the corresponding parallel coordinates representation of the two points.
For simplicity, 
let the distance between the vertical axes $f_1$ and $f_2$ be $1$; 
then it is easy to know that the two lines intersect at a point 
given by $\rho:((1-k)^{-1}, b(1-k)^{-1})$ in parallel coordinates.
This point depends only on $k$ and $b$, 
the parameters of the original line in the Cartesian plane.
This indicates that the parallel coordinates representation of any point on $\ell$ 
passes through the point $\rho$.

%%%% Fig. 16 %%%%

%%%% Fig. 17 %%%%
%%%% Fig. 18 %%%%

From the coordinates of the point $\rho$, 
we can see the relation between the position of $\rho$ and the slope $k$ of the line $\ell$.
If $k<0$,
the intersection occurs between the two parallel coordinates axes.
Especially, 
when $k=-1$ the intersection is precisely midway, 
as in the example of \mbox{Figure~\ref{Fig:MLDMPshapeSDE}(b)}.  
If $1<k$ or $0<k<1$,
then the intersection point is on the left side or right side of the two coordinate axes, 
respectively. 
When $k=\pm\infty$ or $k=0$,
the intersection point is on the left axis or right axis, 
respectively.
Finally, 
when $k=1$,
the lines are parallel between the two axes in parallel coordinates.
The above properties can help us understand the relation between objectives.
For example, 
from the parallel coordinates plot in \mbox{Figure~\ref{Fig:DTLZ5IMshape}} we know that $f_1 = f_2$ and $f_2 = kf_3+b$, 
where $k>1$ and $b=0$.

%%%% Fig. 19 %%%%
\begin{figure}[tbp]
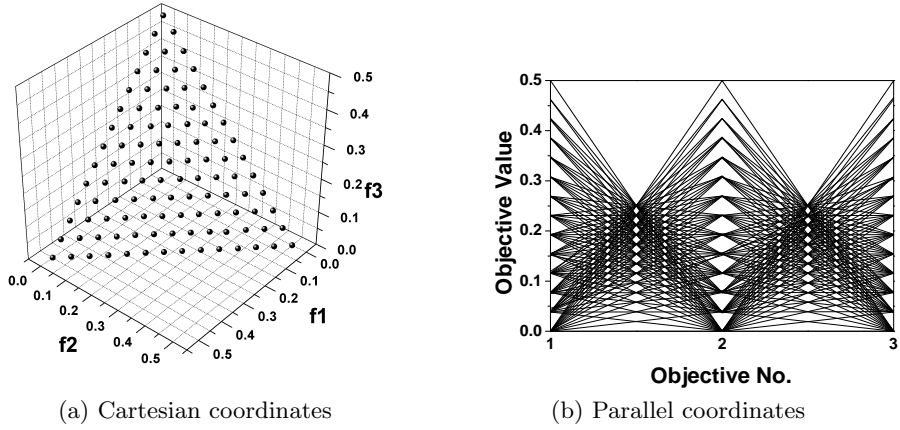

	\begin{center}
		\footnotesize
		\begin{tabular}{@{}cc}
			\includegraphics[scale=0.22]{DTLZ1MOEADCC.pdf}~~~&~~~
			\includegraphics[scale=0.23]{DTLZ1MOEAD.pdf}\\
			(a) Cartesian coordinates & (b) Parallel coordinates \\
		\end{tabular}
	\end{center}
	\vspace{-3mm}
	\caption{The solution set obtained by MOEA/D on the 3-objective DTLZ1.}
	\label{Fig:DTLZ1shapeMOEAD}
\end{figure}

Finally, 
note that since the horizontal position of the intersection point 
in parallel coordinates depends only on the slope $k$, 
when we can see many lines between two objectives in the parallel coordinates plot 
intersecting at same horizontal position (but at different vertical positions), 
this means that many lines connecting two points in the Cartesian coordinate space 
have the same slope with respect to these two objectives.
This occurs often when points in Cartesian coordinates are 
absolutely uniformly-distributed on the plane of these two objectives.
The solution set in \mbox{Figure~\ref{Fig:DTLZ1shapeMOEAD}} has such a pattern 
(see the midway of two adjacent objectives in \mbox{Figure~\ref{Fig:DTLZ1shapeMOEAD}(b)}).
More interesting correspondence between the patterns of lines in parallel coordinates and 
the relation of objectives in the solution set can be found in~\cite{Inselberg2009}.

\section{Objective Order in Parallel Coordinates}

\begin{figure}[tbp]
	\begin{center}
		\footnotesize
		\begin{tabular}{@{}cc}
			\includegraphics[scale=0.22]{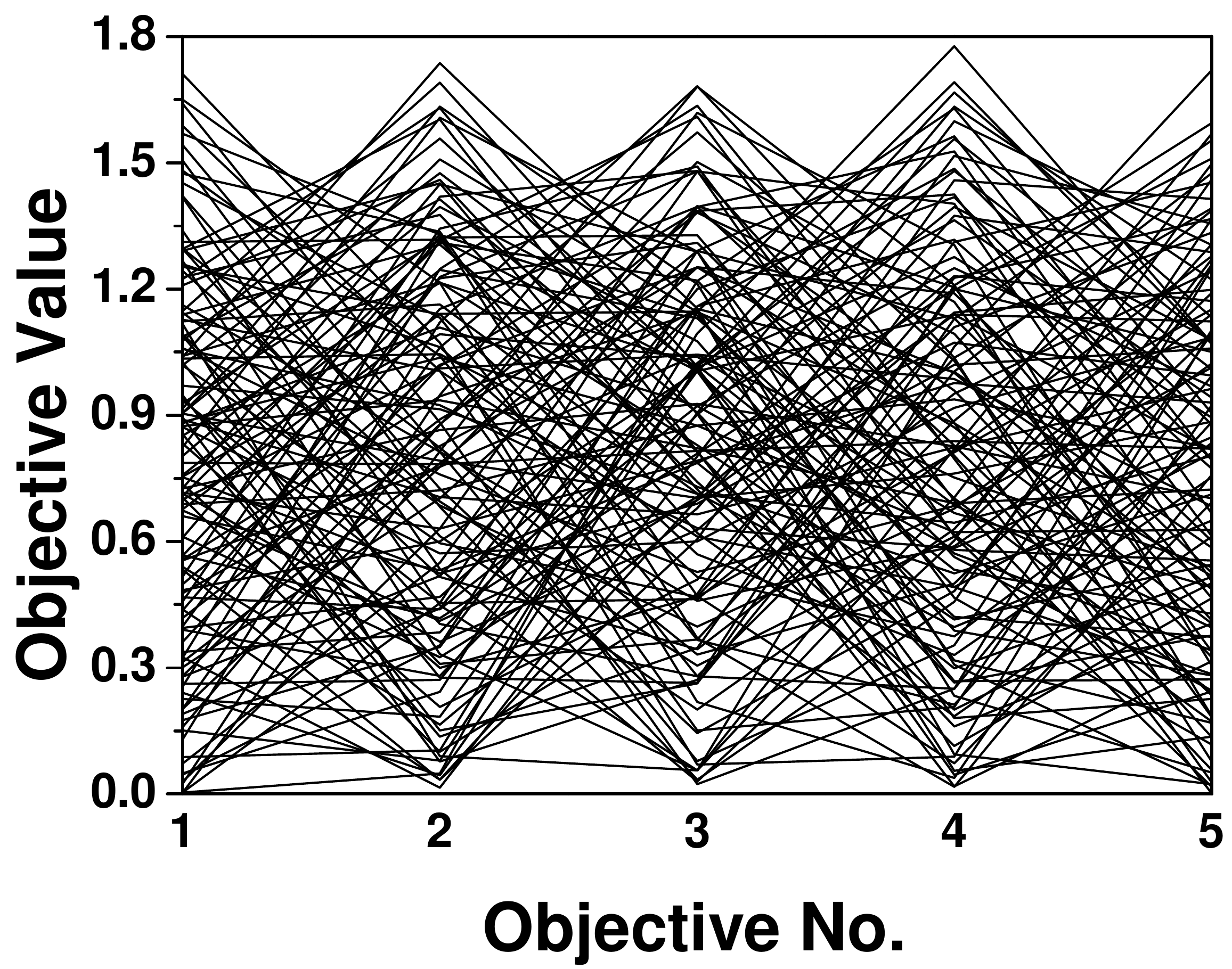}~~~&~~~
			\includegraphics[scale=0.22]{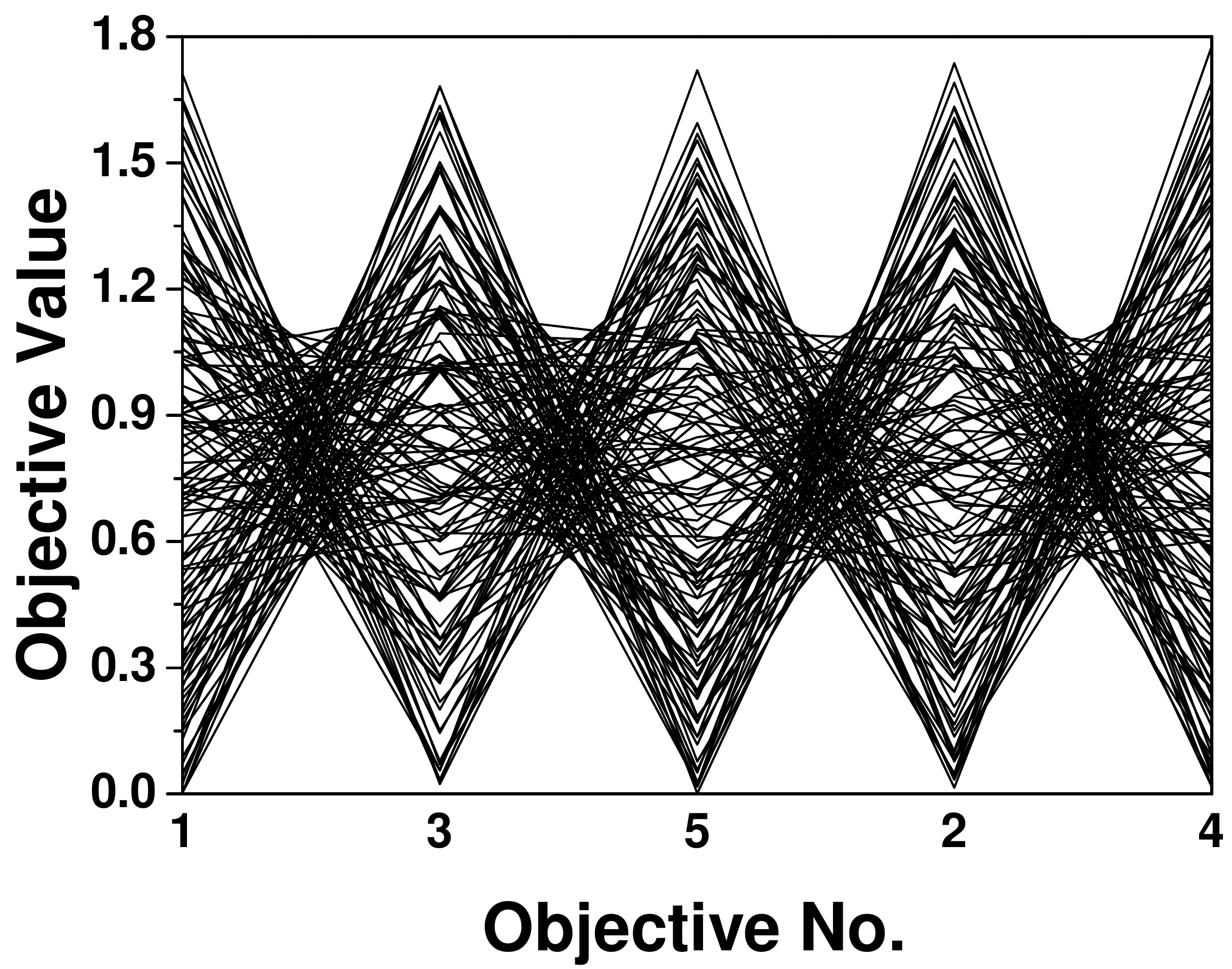}\\
			(a) & (b) \\
		\end{tabular}
	\end{center}
	\vspace{-3mm}
	\caption{The solution set of SPEA2+SDE on the 5-objective ML-DMP, 
		shown by different order of objectives in parallel coordinates.}
	\label{Fig:MLDMPOrder}
\end{figure}

In parallel coordinates, 
each axis has at most two neighboring axes 
(one on the left and one on the right). 
Different order of objective axes presents different information with respect to the relation between objectives. 
Take \mbox{Figure~\ref{Fig:MLDMPOrder}} as an example.
In \mbox{Figure~\ref{Fig:MLDMPOrder}(a)} where the order of objectives is $f_1, f_2, f_3, f_4, f_5$, 
the conflict between any two adjacent objectives is rather weak.
In contrast,
in \mbox{Figure~\ref{Fig:MLDMPOrder}(b)} where the order of objectives is $f_1, f_3, f_5, f_2, f_4$, 
the conflict between any two adjacent objectives is quite intense. 

\begin{figure}[tbp]
	\begin{center}
		\footnotesize
		\begin{tabular}{@{}cc}
			\includegraphics[scale=0.22]{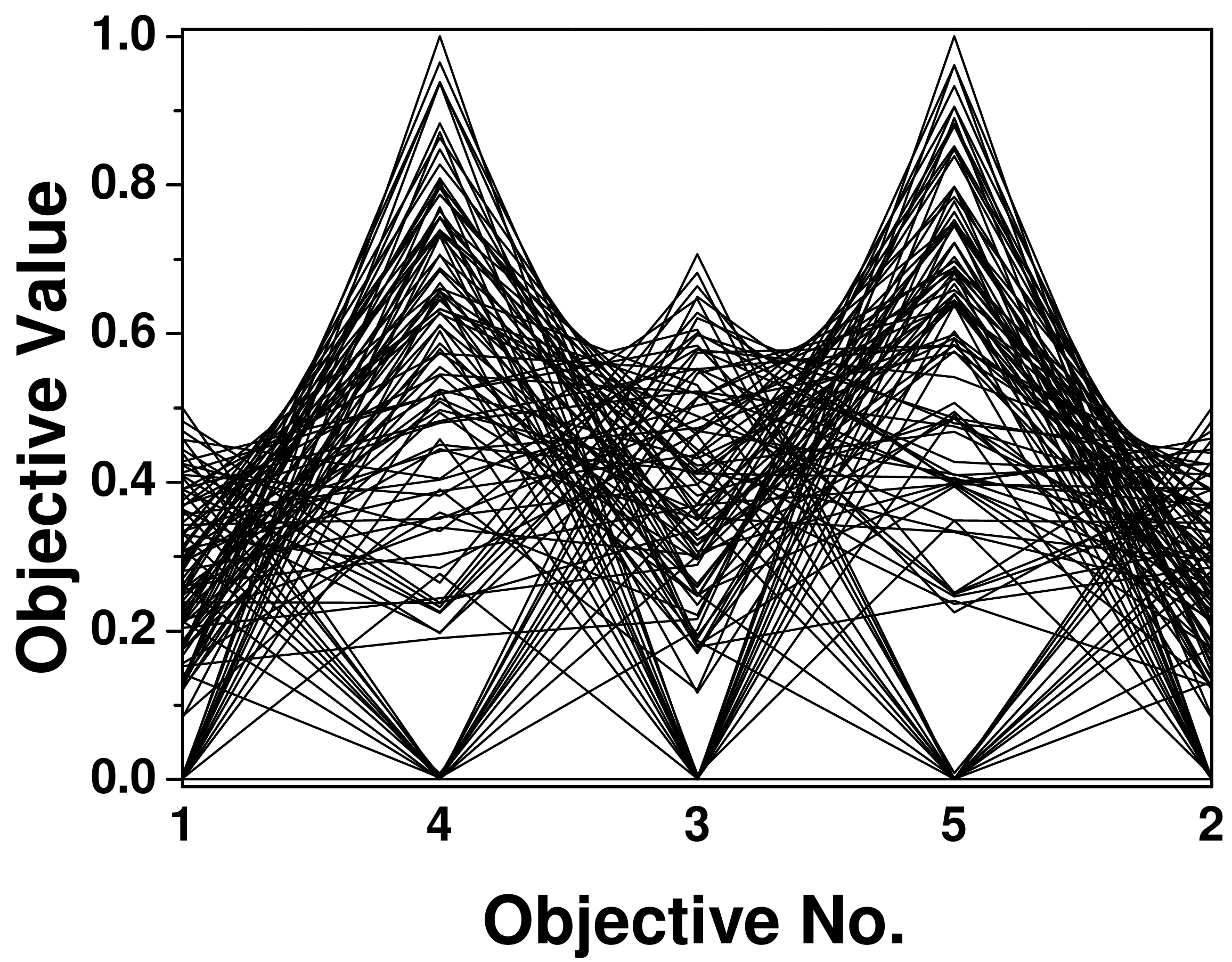}~~~&~~~
			\includegraphics[scale=0.22]{DTLZ5IMshapeSDE.pdf}\\
			(a) & (b) \\
		\end{tabular}
	\end{center}
	\vspace{-3mm}
	\caption{The solution set of SPEA2+SDE on the 5-objective DTLZ5(I,M) where $I=3$, 
		shown by different order of objectives in parallel coordinates.}
	\label{Fig:DTLZ5IMorder}
\end{figure}

\begin{figure}[!]
	\begin{center}
		\footnotesize
		\begin{tabular}{@{}cc}
			\includegraphics[scale=0.22]{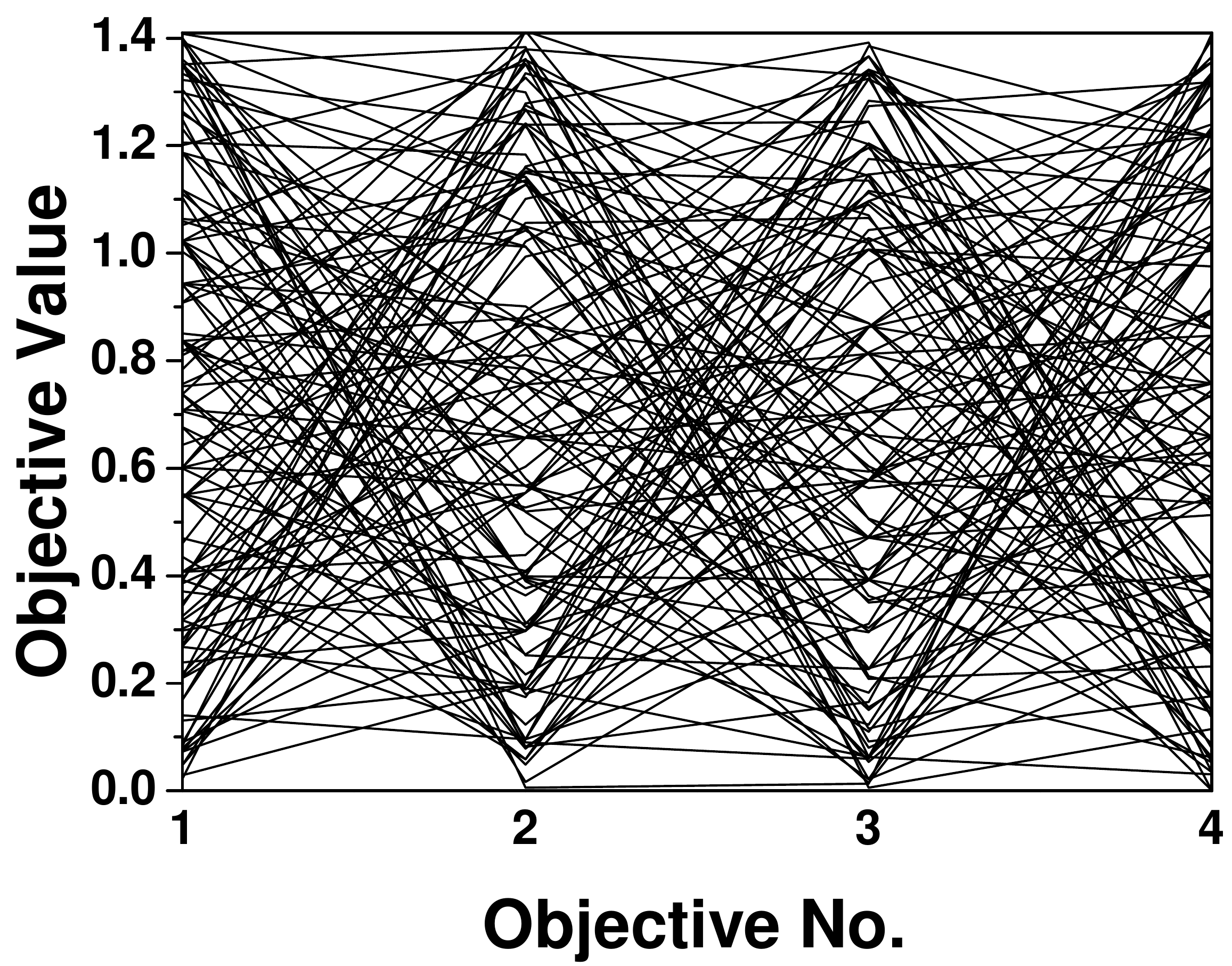}~~~&~~~
			\includegraphics[scale=0.22]{MLDMPshapeSDE.pdf}\\
			(a) & (b) \\
		\end{tabular}
	\end{center}
	\vspace{-3mm}
	\caption{The solution set of SPEA2+SDE on the 4-objective ML-DMP, 
		shown by different order of objectives in parallel coordinates.}
	\label{Fig:MLDMPorder4obj}
\end{figure}

In a solution set with $m$ objectives,
its parallel coordinates representation can only show $m-1$ relationships at a time.
This can be a very small portion compared to the total $\binom{m}{2}$ relationships existing in $m$ objectives.
Therefore,
a good objective axis arrangement providing the user as much (clear) information as possible is of importance.  
As shown in \mbox{Figures~\ref{Fig:DTLZ5IMorder} and \ref{Fig:MLDMPorder4obj}},
after swapping some objectives, 
we can see interesting patterns (linearly dependent) between some pairs of objectives.
However, 
determining a good order of axes is nontrivial.
Researchers in the data visualization area mainly focus on clutter reduction techniques~\cite{Ellis2007, Lu2016}.  
In the multiobjective optimization domain, 
an attempt has been made to place the most harmonious objectives in a row~\cite{De2015}.

%%%% Fig. 20 %%%%
%%%% Fig. 21 %%%%
%%%% Fig. 22 %%%%

\section{How to Draw A Parallel Coordinate Plot}

\begin{figure}[tbp]
	\begin{center}
		\footnotesize
		\begin{tabular}{@{}cccc}
			\includegraphics[width=0.23\textwidth]{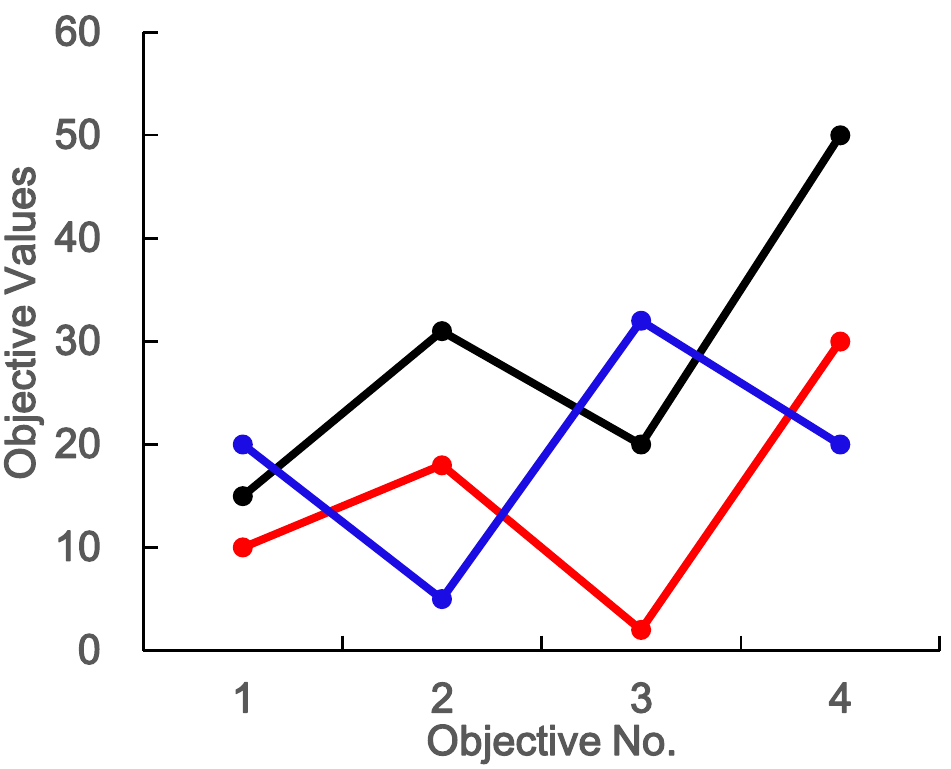}&
			\includegraphics[width=0.23\textwidth]{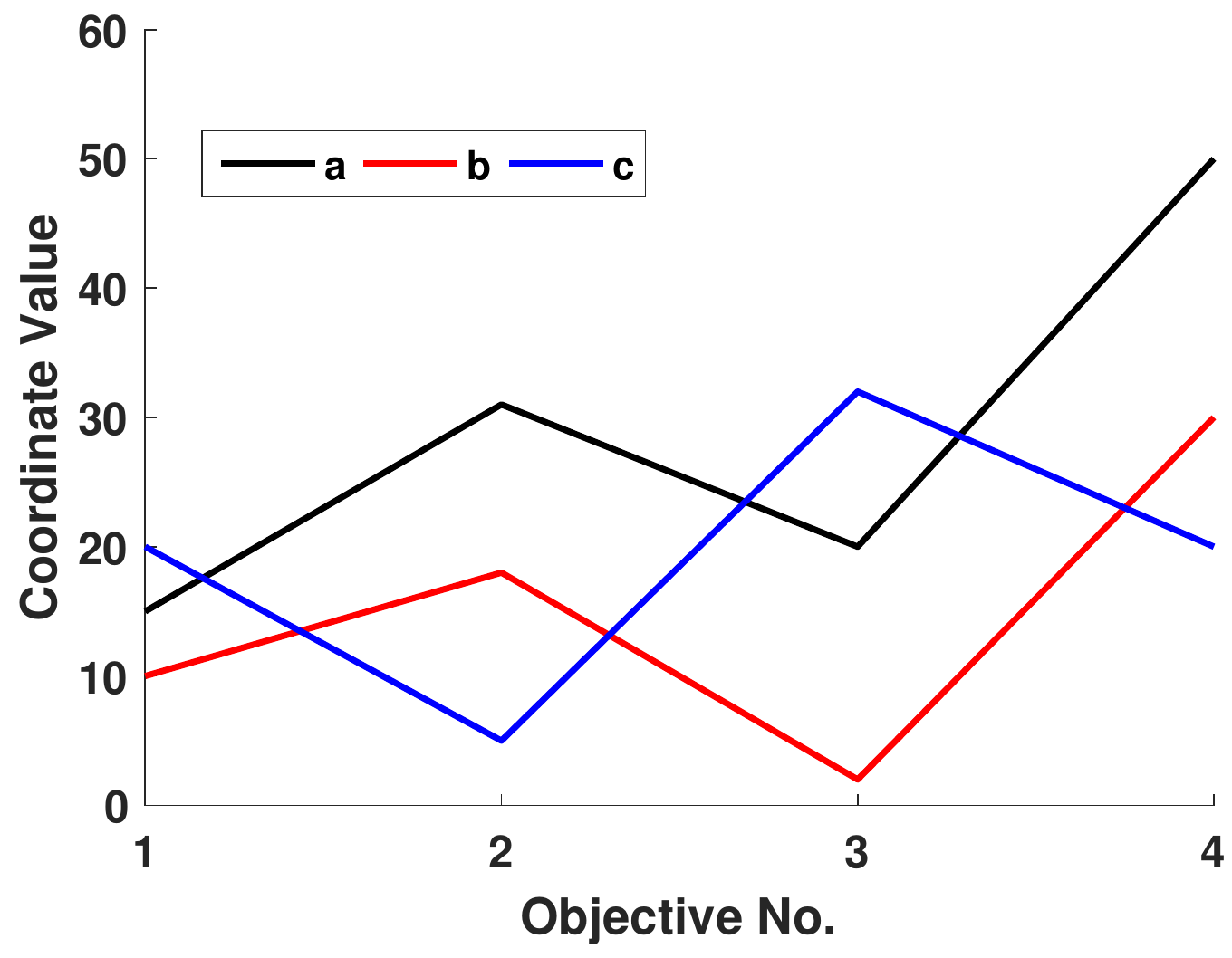}&
			\includegraphics[width=0.20\textwidth]{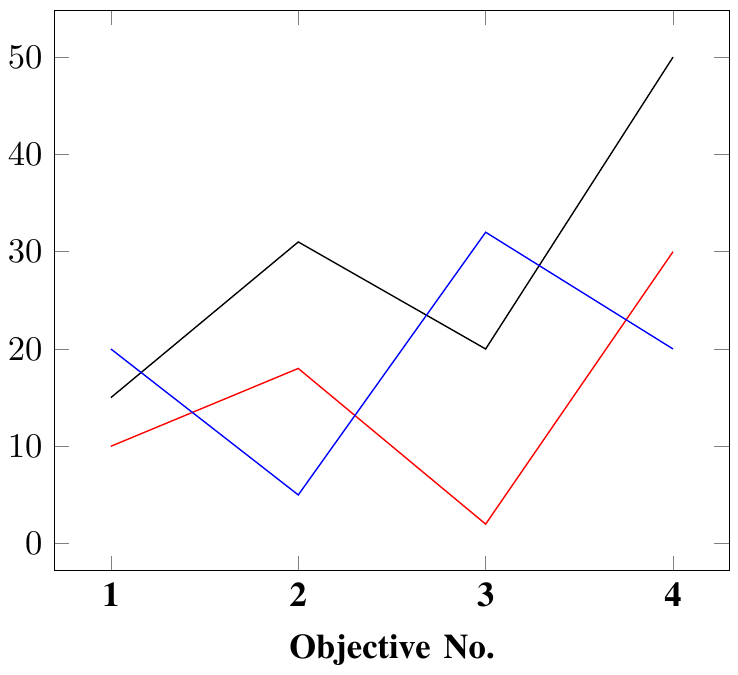}&
			\includegraphics[width=0.24\textwidth]{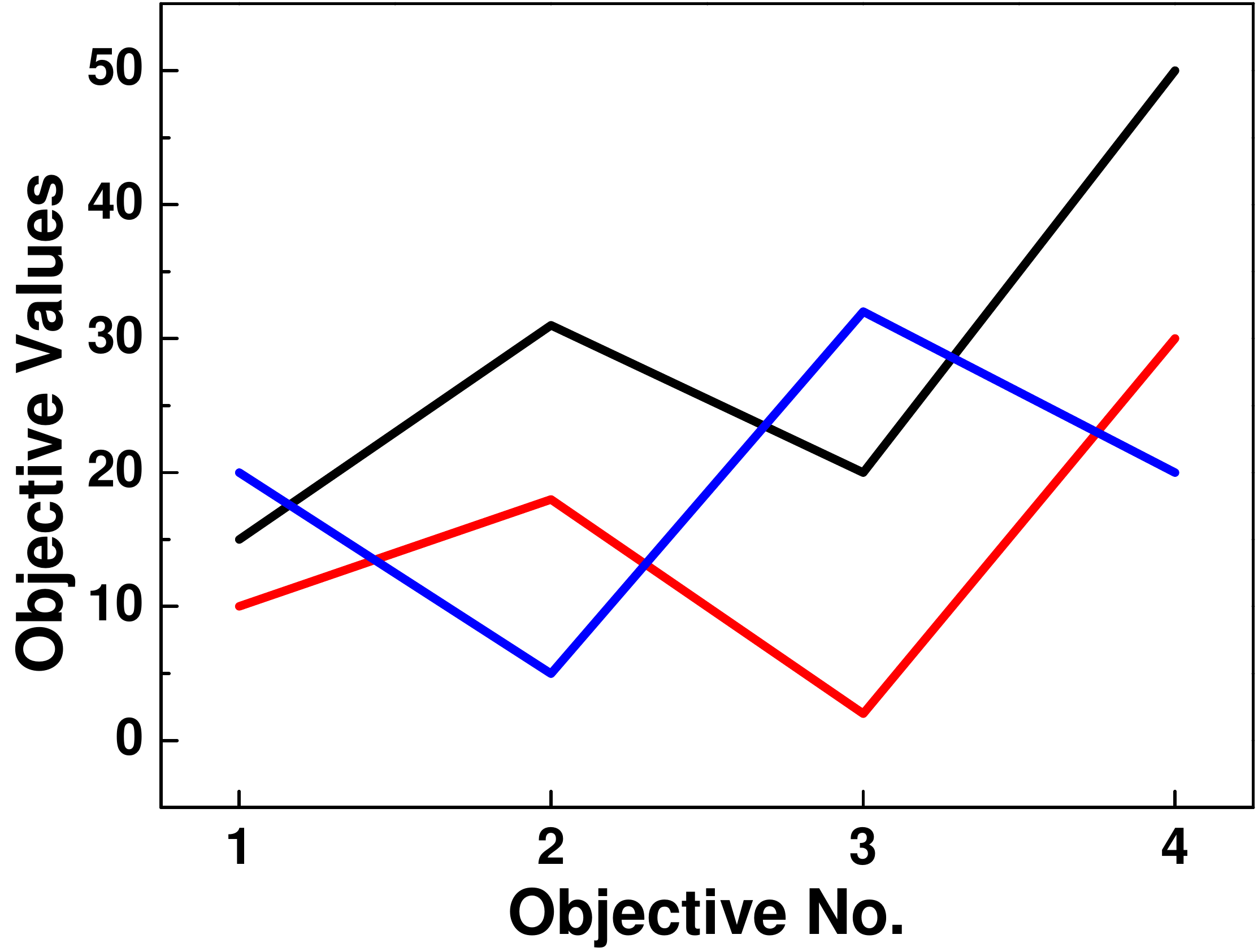}\\
			(a) MS Excel & (b) MATLAB & (c) Latex & (d) Origin\\
		\end{tabular}
	\end{center}
	\vspace{-3mm}
	\caption{An example of parallel coordinates plots in MS Excel, MATLAB, Latex, and Origin.}
	\label{Fig:PCplotting}
\end{figure}

In this section,
we give procedures of how to plot a solution set in parallel coordinates by several commonly-used graphing tools: 
MS Excel, MATLAB, Latex, and Origin. 
Tables~\ref{Table:Excel}--\ref{Table:Origin} provide the steps/codes by MS Excel, MATLAB, Latex, and Origin, 
respectively.
For the command-line format tools (i.e., MATLAB and Latex), 
we use the example in Figure~\ref{Fig:PC}.
Figure~\ref{Fig:PCplotting} also presents the graphs drawn by the four tools to that example.
Finally, 
we would like to note that in this paper all of the parallel coordinates graphs of data examples were drawn by Origin.

\section{Conclusions}

Parallel coordinates have drawn increasing attention in many-objective optimization,
but mapping a many-objective solution set onto a 2D parallel coordinates plane may not be straightforward 
to reveal the information contained in the set.
This paper has made some observations on the use of parallel coordinates 
to present a solution set in many-objective optimization.
In particular,
\begin{itemize}
	\item the parallel coordinates representation of a solution set can \textit{partly} reflect its convergence, 
	coverage and uniformity. 
	This suggests that parallel coordinates can be an assistant tool 
	(but not entirely replacing quality metrics) 
	in assessing a many-objective solution set.

	\item although the clarity can be affected by overlapping polygonal lines,
	parallel coordinates transform certain geometrical features of a many-objective solution set 
	into easily seen 2D patterns.

	\item the order of objective axes matters in parallel coordinates. 
	To better present the relationship between objectives,
	it may need to be rearranged according to features of the solution set considered.
	
\end{itemize}

\begin{table}[tbp]\normalsize
	\caption{Steps of creating a parallel coordinates plot in MS Excel.}
	\label{Table:Excel}
	\centering
	\begin{tabular}{|p{8 cm} |}
		\hline
		\\
		(1) Input the data as a table with each row as a solution and select them.\\
		(2) Click Insert -$>$ Recommended Charts.\\
		(3) On the Recommended Charts tab, scroll through the list of charts that Excel recommends for your data, click Line chart -$>$ OK.\\
		(4) Use the Chart Elements, Chart Styles, and Chart Filters buttons next to the upper-right corner of the chart to add chart elements like axis titles or data labels, customize the look of your chart, or change the data shown in the chart.\\\\
		\hline
	\end{tabular}
\end{table}
\begin{table}[!]\normalsize
	\caption{Codes of creating a parallel coordinates plot in MATLAB.}
	\label{Table:Matlab}
	\centering
	\begin{tabular}{|p{8 cm} |}
		\hline\\
		X = [15 31 20 50; 10 18 2 30; 20 5 32 20];\\
		groups = \{'a', 'b', 'c'\};\\
		parallelcoords(X,'group',groups);\\
		xlabel('Objective No.');\\\\
		\hline
	\end{tabular}
\end{table}
\begin{table}[!]\normalsize
	\label{Table:Latex}
	\caption{Codes of creating a parallel coordinates plot in Latex using the PGFPlots package.}
	\centering
	\begin{tabular}{|p{8 cm} |}
		\hline\\
		$\backslash$usepackage\{pgfplots\}\\
		$\backslash$begin\{tikzpicture\}\\
		$\backslash$begin\{axis\}[xlabel=\{Objective No.\},symbolic x coords=\{1, 2, 3, 4\},xtick=data]\\
		$\backslash$addplot+[mark=none,draw=green,sharp plot] plot coordinates \{(1,15) (2,31) (3,20) (4,50)\};\\
		$\backslash$addplot+[mark=none,draw=blue,sharp plot] plot coordinates \{(1,10) (2,18) (3,2) (4,30)\};\\
		$\backslash$addplot+[mark=none,draw=red,sharp plot] plot coordinates \{(1,20) (2,5) (3,32) (4,20)\};\\
		$\backslash$end\{axis\}\\
		$\backslash$end\{tikzpicture\}\\\\
		\hline
	\end{tabular}
\end{table}
\begin{table}[!]\normalsize
	\caption{Steps of creating a parallel coordinates plot in Origin.}
	\label{Table:Origin}
	\centering
	\begin{tabular}{|p{8 cm} |}
		\hline
		\\
		(1) Create a table consisting of the first column being X axis from $1$ to $m$ (where $m$ is the number of objectives) 
		and the remaining columns being Y axis with each column for a solution. \\ 
		(2) Select the table and click the Line button at the lower-left corner of the panel. \\ 
		(3) Double click the Axis Labels and the polylines of parallel coordinates to customize the look of the chart. \\\\
		\hline
	\end{tabular}
\end{table}

%%%% Fig. 23 %%%%

Our subsequent study is towards overcoming/alleviating the difficulties of 
interpreting parallel coordinates plots presented in this paper.
Particularly,
how to arrange the order of objectives will be the focus of our future work 
as it had presented its usefulness in the paper.
In this regard, 
a straightforward thought is to place the most conflicting objectives 
or the most harmonious objectives together 
so that people could see some meaningful patterns (such as the examples 
in \mbox{Figures~\ref{Fig:DTLZ5IMorder} and~\ref{Fig:MLDMPorder4obj}}).
Another thought is to consider the coverage of lines between objectives in a parallel coordinates plot;
people may acquire more information from less coverage of the lines, 
for example, 
after exchanging the order of objectives $f_1$ and $f_2$ in \mbox{Figure~\ref{Fig:ExampleCoverage}}.

\section*{Acknowledgement}
This work was supported in part by the Engineering and Physical
Sciences Research Council (EPSRC) of U.K. under Grants EP/K001523/1 and EP/J017515/1, 
and National Natural Science Foundation of China (NSFC) under Grants 61329302 and 61403326.
X.~Yao was also supported by a Royal Society Wolfson Research Merit Award.

%\footnotesize
%\bibliographystyle{unsrt}
%\bibliographystyle{IEEEtranS}
%\bibliographystyle{IEEEtran}
%\bibliography{IEEEabrv,bibfile}

\end{document}